\documentclass[table]{article}

\usepackage[letterpaper,top=2cm,bottom=2cm,left=3cm,right=3cm,marginparwidth=1.75cm]{geometry}

\usepackage{amsmath}
\usepackage{graphicx}
\usepackage[colorlinks=true, allcolors=blue]{hyperref}
\usepackage{authblk}
\usepackage{float}
\usepackage{pdfpages}

\usepackage{booktabs}
\usepackage{multirow}
\usepackage{amssymb}
\makeatletter
\def\hlinewd#1{%
\noalign{\ifnum0=`}\fi\hrule \@height #1 %
\futurelet\reserved@a\@xhline}
\makeatother
\usepackage{diagbox}
\newcommand{\remove}[1]{}
\usepackage{xcolor}
\usepackage{subcaption}
\usepackage{fvextra}
\usepackage[numbers]{natbib}

\title{GneissWeb: Preparing High Quality Data for LLMs at Scale}
\author[*]{Hajar Emami Gohari}
\author[*]{Swanand Ravindra Kadhe}
\author[ \!\!]{Syed Yousaf Shah$^\dag$}
\author[ \!\!]{Constantin Adam}
\author[ \!\!]{Abdulhamid Adebayo}
\author[ \!\!]{Praneet Adusumilli}
\author[ \!\!]{Farhan Ahmed}
\author[ \!\!]{Nathalie Baracaldo Angel}
\author[ \!\!]{Santosh Subhashrao Borse}
\author[ \!\!]{Yuan-Chi Chang}
\author[ \!\!]{Xuan-Hong Dang}
\author[ \!\!]{Nirmit Desai}
\author[ \!\!]{Revital Eres}
\author[ \!\!]{Ran Iwamoto}
\author[ \!\!]{Alexei Karve}
\author[ \!\!]{Yan Koyfman}
\author[ \!\!]{Wei-Han Lee}
\author[ \!\!]{Changchang Liu}
\author[ \!\!]{Boris Lublinsky}
\author[ \!\!]{Takuyo Ohko}
\author[ \!\!]{Pablo Pesce}
\author[ \!\!]{Maroun Touma}
\author[ \!\!]{Shiqiang Wang}
\author[ \!\!]{Shalisha Witherspoon}
\author[ \!\!] {Herbert Woisetschl\"ager}
\author[ \!\!]{David Wood}
\author[ \!\!]{Kun-Lung Wu}
\author[ \!\!]{Issei Yoshida}
\author[ \!\!]{Syed Zawad}
\author[ \!\!]{Petros Zerfos}
\author[ \!\!]{Yi Zhou}
\author[ \!\!]{Bishwaranjan Bhattacharjee$^\dag$}

\affil[ ] {IBM Research}
\affil[ ] {* Equal Contribution}
\affil[ ] {\dag  Corresponding Authors}
\affil[ ] {syshah@us.ibm.com, bhatta@us.ibm.com}

\newcommand{\myparagraph}[1]{\medskip\noindent\textbf{#1}}

\newcommand{\readability}{\mathsf{Readability}}

\newcommand{\sciClassifier}{\phi_{\textrm{sci}}}
\newcommand{\eduClassifier}{\phi_{\textrm{edu}}}
\newcommand{\techClassifier}{\phi_{\textrm{tech}}}
\newcommand{\medClassifier}{\phi_{\textrm{med}}}
\newcommand{\fasttexDCLM}{\phi_\textrm{DCLM}}
\newcommand{\fasttexCosmo}{\phi_\textrm{Cosmo}}
\newcommand{\thresholdDCLM}{\tau_\textrm{DCLM}}
\newcommand{\thresholdCosmo}{\tau_\textrm{Cosmo}}
\newcommand{\thresholdRscoreCat}{r_c}
\newcommand{\thresholdETCatLow}{\tau^{\textrm{Low}}_c}
\newcommand{\thresholdETCatHigh}{\tau^{\textrm{High}}_c}

\begin{document}
\maketitle

\begin{abstract}
Data quantity and quality play a vital role in determining the performance of Large Language Models (LLMs). High-quality data, in particular, can significantly boost the LLM's ability to generalize on a wide range of downstream tasks. Large pre-training datasets for leading LLMs remain inaccessible to the public, whereas many open datasets are small in size (less than 5 trillion tokens), limiting their suitability for training large models. 


In this paper, we introduce GneissWeb, a large dataset yielding around 10 trillion tokens that caters to the data quality and quantity requirements of training LLMs. Our GneissWeb recipe that produced the dataset consists of sharded exact sub-string deduplication and a judiciously constructed ensemble of quality filters. GneissWeb achieves a favorable trade-off between data quality and quantity, producing models that outperform models trained on state-of-the-art open large datasets (5+ trillion tokens). 
We show that models trained using GneissWeb dataset outperform those trained on FineWeb-V1.1.0 by 2.73 percentage points in terms of average score computed on a set of 11 commonly used benchmarks (both zero-shot and few-shot) for pre-training dataset evaluation. When the evaluation set is extended to 20 benchmarks (both zero-shot and few-shot), models trained using GneissWeb still achieve a 1.75 percentage points advantage over those trained on FineWeb-V1.1.0.
\end{abstract}

\section{Introduction}
\label{intro} 

Large Language Models (LLM) are becoming pervasive in many aspects of life. The performance of these models are dictated by several factors including the model architecture, model size, training data size as well as training data quality.  

How much data should one use to train an LLM of certain size? The answer is typically governed by scaling laws -- empirical formulas that estimate optimal models sizes and data sizes for a given compute budget. For instance, the widely adopted \textit{Chinchilla} law~\cite{hoffman2022training} suggested a compute optimal token-to-parameter-ratio of roughly 20. However, recent state-of-the-art LLMs have been trained on far more data than what the scaling laws would deem as optimal. For instance, Llama3 family of models are trained on 15 trillion (15T) tokens (compared to 1.8T tokens for Llama2) \cite{grattafiori2024llama3,touvron2023llama2}, Gemma2 family of models are trained on 13T tokens \cite{gemmateam2024gemma2}, and Granite-3.0 family of models are trained on 12T tokens \cite{granite30}. At the time of writing of this paper, the pre-training datasets for leading LLMs, such as Llama3 \cite{grattafiori2024llama3} and Mixtral \cite{jiang2024mixtral}, remain inaccessible to the public, with limited information available on their creation process. 

Opacity of the datasets used to train leading LLMs, has motivated the development of several open-source datasets \cite{penedo2023refinedweb,cerebras2023slimpajama,weber2024redpajama,soldaini2024dolma,li2024datacomplm}.
These datasets are mainly derived by processing text from the Common Crawl \cite{commoncrawl} and optionally mixing some high-quality data sources (e.g., GitHub). However, majority of these datasets are less than 5T tokens which limits their suitability for pre-training large LLMs. In particular, large LLMs typically undergo long token horizon pre-training consisting of two stages \cite{granite30}. In Stage-1 of pre-training, the model is trained on a very large corpus of data to cover the breadth, followed by a Stage-2 pre-training which uses much higher quality but comparatively smaller dataset to further improve the model. Data quantity and quality play a crucial role in determining the performance of LLMs. High-quality data significantly boosts the LLM's ability to generalize on a wide range of downstream tasks, making it cheaper to train better models. This delicate interplay between data quality and quantity makes it challenging to develop large-scale, high-quality pre-training datasets that are suitable to  Stage-1 long token horizon training.

In this paper, we introduce \textbf{GneissWeb}\footnote{\textit{Gneiss}, pronounced “nice”, is a durable igneous rock, just like IBM’s open-source Granite models trained from it..} dataset along with the recipe of how we produced this dataset. The GneissWeb recipe consists of sharded exact substring deduplication and a judiciously constructed ensemble of quality filters. 
The GneissWeb recipe is built by developing novel processing steps and quality filters that can effectively identify and filter out low-quality data. We go beyond simple model-based quality filtering used in recent datasets and design an ensemble of filters incorporating novel quality filters based on characteristics of the text contents. We develop a novel quality filtering called ``Extreme Tokenized Documents Removal'' that effectively leverages information from both the ``pre-tokenization'' stage and the ``post-tokenization'' stage to filter out low-quality documents based on tokenized data that is used in LLM training. Our novel readability score quality filter is another innovative processing step that effectively utilizes information based on human ability of reading documents from different domains for identifying and excluding low-quality documents. Furthermore, we leverage the domain information as category of a document in our quality filtering process which reduces the risk of loosing high-quality data by processing all documents in the same way.

We design the GneissWeb recipe thoroughly analyzing and testing each key ingredient implemented in GneissWeb recipes, conducting comprehensive evaluations of various quality filtering approaches and deduplication methods. We present the key evaluations that guided our design choices and provide filtering thresholds that can be used to filter the dataset, to match the token quantity and quality needs of Stage-1. To cater to the long horizon training needs of LLMs, we focused on the goal to produce a dataset that can generate $\sim$10T tokens that are higher quality than all other open datasets of similar size. GneissWeb is fully prepared using our publicly released IBM data-prep-kit\footnote{https://github.com/IBM/data-prep-kit}, with the majority of data preparation steps efficiently running at scale on Kubernetes clusters.

\begin{figure*}[t!]
    \centering
    \begin{subfigure}[t]{0.5\textwidth}
        \centering
        \includegraphics[scale=0.525]{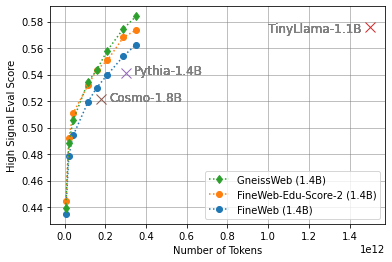}
    \end{subfigure}%
    ~ 
    \begin{subfigure}[t]{0.5\textwidth}
        \centering
        \includegraphics[scale=0.525]{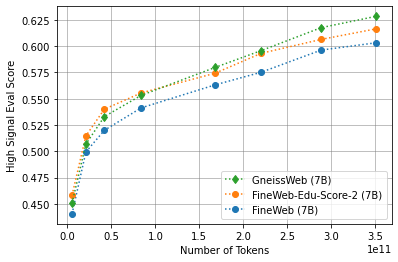}
    \end{subfigure}
    \caption{\textbf{GneissWeb ($\sim$10T tokens) outperforms state-of-the-art open-source datasets with 5T+ tokens.} Specifically, we compare average scores on a set of 11 tasks with 18 variants (zero-shot and few-shot) for 1.4B parameter models (left) and 7B parameter models (right), trained on 350B tokens. We also compare with state-of-the-art existing models of roughly 1B parameter size. Models trained on GneissWeb (green) achieve higher performance than the models trained on other datasets (circles) and existing models (crosses).}
    \label{fig:gneissweb_vs_others_intro}
\end{figure*}

Our evaluations demonstrate that GneissWeb outperforms state-of-the-art large open datasets (5T+ tokens). Specifically, 1.4B parameter models trained on GneissWeb outperform those trained on FineWeb-V1.1.0 \cite{penedo2024fineweb} by 2.73 percent points in terms of average score computed on a set of 11 benchmarks (both zero-shot and few-shot) commonly used to evaluate pre-train datasets. When the evaluation set is extended to 20 benchmarks (both zero-shot and few-shot), models trained on GneissWeb outperform those trained on FineWeb-V1.1.0 by 1.75 percent points. GneissWeb also produces better performing models at the 3B and 7B model sizes compared state-of-the-art large open datasets (5T+ tokens). See Figure \ref{fig:gneissweb_vs_others_intro} for summary of key evaluation results and Section \ref{sec:evaluation} for details.

\section{Related Work} 
\label{relatedwork}


In this work we aim to create a large dataset capable for pre-training of a LLM. There are several related works in this space. Prior public pre-training datasets are typically derived from the Common Crawl ~\cite{commoncrawl}. Early works include the C4 dataset with 160 billion tokens \cite{raffel2020exploring} and the Pile dataset with billion tokens \cite{gao2020pile}. The C4 dataset is curated from the April 2009 snapshot of the Common Crawl. It uses langdetect \cite{langdetect} to detect English text, applies a series of heuristic filters including discarding any page with less than 3 sentences, removing lines without any terminal punctuation mark, removing any page containing any word in a list of dirty, naughty, obscene or bad words etc, and also performs deduplication by removing all but one of any three-sentence span occurring more than once in the dataset. The Pile is a composite dataset that includes the Pile-CC, which is based on Common Crawl. It uses pycld2 \citep{pycld2} for language detection, removes boilerplate using jusText \citep{justext}, applies classifier-based filtering and performs fuzzy deduplication. 

Multilingual models like XLM RoBERTa \cite{XLMRoBERTa} used the CC100 dataset \cite{conneu2019cc100}. This dataset was curated using the CCNET \cite{wenzek2019ccnet} processing pipeline on one year of Common Crawl snapshots. CCNet uses the data processing methods introduced in fastText \cite{joulin2017bag}, which include deduplicating documents and applying LangID filtering. It then adds a filtering step to select documents that are similar to high-quality corpora like Wikipedia by utilizing a 5-gram KenLM filter.

RedPajama dataset \cite{weber2024redpajama} is an open source attempt to recreate the dataset used to train Llama models. It is a composite dataset which includes text obtained from the Common Crawl by using the CCNet pipeline \cite{wenzek2019ccnet} and a classifier trained to identify documents similar to Wikipedia articles or references. SlimPajama with 627B tokens \cite{cerebras2023slimpajama} further refines RedPajama by removing short documents and performing additional fuzzy dedupllication. RedPajama-V2 \cite{weber2024redpajama} with 30 trillion tokens is entirely based on the Common Crawl and contains annotations without applying any filtering. These annotations cover filtering techniques from CCNet, C4, and others, and also labels identifying deduplicates using exact and fuzzy deduplication. 

RefinedWeb dataset \cite{penedo2023refinedweb} is a Common Crawl-based dataset, using trafilatura \cite{trafilatura} for text extraction, fastText-based language identification \cite{joulin2017bag}, heuristic rules for quality filtering, and fuzzy and exact deduplication. Dolma \cite{soldaini2024dolma} is a 3 trillion token composite dataset with a Common Crawl-based portion, which employs fastText for language identification, primarily uses heuristic rules from MassiveWeb \cite{rae2021scaling} for quality filtering, applies toxicity filtering based on rules and classifiers and performs deduplication at URL, document and paragraph levels. 

More recent datasets include FineWeb datasets \cite{penedo2024fineweb}, DCLM-Baseline \cite{li2024datacomplm}, and TxT360 \cite{txt360data2024}. FineWeb consists of 15T tokens derived from the Common Crawl by applying a series of processing steps, mainly including language classification, fuzzy deduplication at snapshot level and heuristic rule-based quality filters. 
Subsequently, two smaller but higher quality versions called FineWeb-Edu (1.3 trillion tokens) and FineWeb-Edu-Score2 (5.4 trillion tokens) derived from FineWeb were released \cite{penedo2024fineweb}. These smaller high quality derivatives of FineWeb are created by retaining documents perceived to have higher educational value from FineWeb. See Appendix \ref{app:fineweb} for more details on FineWeb.

DCLM-Baseline is obtained from the Common Crawl snapshots by using resiliparse \cite{resiliparse} for text extraction,  heuristic quality filters from RefinedWeb, fuzzy deduplication with Bloom filter \cite{bff}, model-based quality filtering using a specially trained fastText classifier. TxT360 is a composite dataset obtained from Common Crawl snapshots and 14 high-quality datasets (e.g. FreeLaw, Ubuntu IRC, etc). TxT360 is obtained by first applying local exact deduplication, global fuzzy deduplication, and quality filtering to both web and curated datasets, resulting in approximately 5 trillion tokens, which are then up-sampled to over 15 trillion tokens. The mixing and up-sampling approach is shown essential to boosting TxT360 performance. 

Nemotron-CC \cite{su2024nemotron} and Zyda2 \cite{tokpanov2024zyda} are concurrent works published recently. Zyda-2 is a 5 trillion high-quality token dataset obtained by collating high-quality open-source datasets including FineWeb-Edu, DCLM, Zyda-1, and Dolma-CC and then applying cross-deduplication and model-based quality filtering. Nemotron-CC is a 6.3 trillion token dataset, including 4.4 trillion tokens from Common Crawl by applying exact substring deduplication, global fuzzy deduplication and model-based quality filtering. Nemotron-CC also includes 1.9 trillion synthetic tokens (approximately 30\% of the data) generated using a rephrasing-based approach from low-quality and high-quality documents.

We take FineWeb \cite{penedo2024fineweb} as the starting point to build our dataset since FineWeb is sufficiently large dataset with 15T tokens which has been shown to outperform several public datasets -- C4, RefinedWeb, Dolma, RedPajamaV, SlimPajama and the Pile.
While FineWeb-Edu, FineWeb-Edu-Score-2 \cite{penedo2024fineweb} and the recent DCLM-Baseline \cite{li2024datacomplm} improve data quality over FineWeb they do so by performing aggressive model-based quality filtering. Such an aggressive filtering cuts down their size which may not be sufficient for pre-training (as pre-training typically consists of only one pass or few passes over the pre-training dataset \cite{muennighoff2023scaling}). Our GneissWeb recipe achieves a favorable trade-off between data quality and quantity thereby producing $\sim$10T high quality tokens with higher performance than prior datasets with 5T+ tokens.

\section{GneissWeb Dataset in a Nutshell}
\label{sec:gneissweb}

\begin{figure}[h!]
    \centering
    \includegraphics[width=0.9\linewidth]{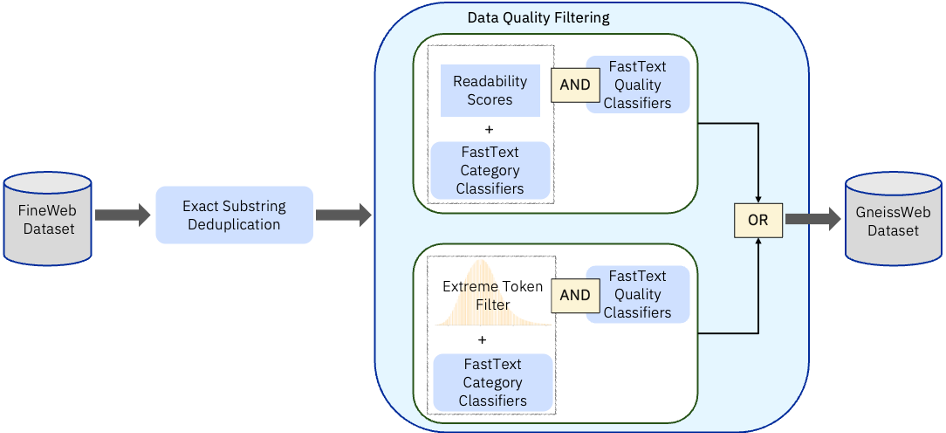}
    \caption{An Outline of the GneissWeb recipe.}
    \label{fig:GneissWeb}
\end{figure}



\noindent\textbf{Building on Top of FineWeb:} We use FineWeb-V1.1.0 as base dataset for GneissWeb, with the goal of obtaining sufficiently large number of quality tokens that are suitable for Stage-1 pre-training. We developed the \textit{GneissWeb recipe} to distill $\sim$10T high quality tokens from FineWeb. We produced the GneissWeb dataset with nearly 10T tokens by applying the GneissWeb recipe to the 15T tokens of FineWeb-V1.1.0, however, FineWeb dataset is not a requirement for our \textit{GneissWeb recipe} neither is it tied to FineWeb.

A key differentiator of the GneissWeb recipe is that it employs a multi-faceted ensemble of quality annotators and thresholds can be adjusted at annotator level to filter documents based on use-case. This is in contrast with recent high-quality datasets  \cite{penedo2024fineweb,li2024datacomplm}, which rely on a single model-based quality annotator and perform aggressive filtering which removes around 90\% of data. Such aggressive filtering, although improves data quality, results in substantial reduction in data quantity and limits the applicability of these datasets for Stage-1 pre-training. The ensemble of quality annotators in the GneissWeb recipe enables fine-grained quality filtering and achieves a favorable trade-off between the data quality and quantity.

We note that, while the GneissWeb recipe is focused at obtaining nearly 10T high quality tokens suitable for Stage-1 pre-training, it is also possible to adapt the recipe by tuning filtering parameters to produce smaller and higher quality datasets fit for Stage-2 type of pre-training.



\vspace{2pt}
\noindent\textbf{The GneissWeb Recipe}
consists of the following ingredients:
\begin{itemize}
\item Exact substring deduplication at line level (Sec. \ref{sec:substring_dedup})
\item Ensemble quality filter (Sec. \ref{sec:ensemble_filter}) consisting of 
\begin{itemize}
\item Custom built combination of fastText Classifiers (Sec. \ref{sec:fastText})
\item Custom built fastText Category Classifiers (Sec. \ref{sec:category_classifiers})
\item Custom built Category-Aware Readability Score Filter (Sec. \ref{sec:readability})
\item Custom built Category-Aware Extreme-Tokenized-Documents Filter (Sec. \ref{sec:extreme_tokenized})
\end{itemize}
\end{itemize}

There are various ways to combine the key ingredients and build a recipe, including deciding which components to include and their order as well as designing ensemble filtering rules using multiple quality annotators. 
We performed rigorous ablations by combining the key ingredients in multiple variations and sequences with the aim of maximizing downstream task performance under the constraint of retaining at least 10T tokens from FineWeb.V1.1.0 (Sec. \ref{sec:ablations}). 
The GneissWeb recipe illustrated in Figure \ref{fig:GneissWeb} produces the highest performance gain. Applying the GneissWeb recipe to 15T tokens of FineWeb-V1.1.0 produces the GneissWeb dataset with 10T tokens.


\section{The GneissWeb Recipe}

In this section we provide details of individual components of the GneissWeb recipe.  
\begin{figure*}[h!]
\centering
  \includegraphics[width=0.5 \textwidth]{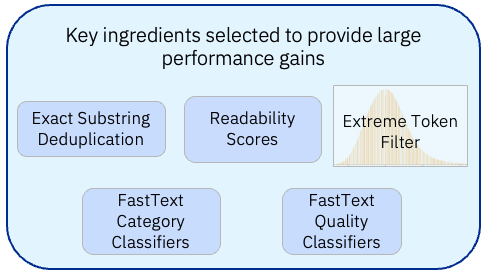}
  \caption{Key ingredients selected for building the GneissWeb recipe.}
  \label{fig:Key}
\end{figure*}

\subsection{Exact Substring Deduplication}
\label{sec:substring_dedup}
Removing duplicates from training data has been shown to reduce memorization \cite{kandpal2022deduplicating,carlini2023quantifying} and improve model performance \cite{lee2021deduplicating,penedo2023refinedweb}. 
FineWeb applied per snapshot fuzzy deduplication and removed near-duplicate documents using the MinHash algorithm \cite{penedo2024fineweb}. 
Furthermore, FineWeb also applied repetition filter, intra-document deduplication \cite{rae2021scaling} which removes documents with many repeated lines and paragraphs. (See Appendix \ref{app:fineweb} for details.)
However, duplicates still remain at sequence-level within and across documents. Such repeated substrings bypass the \textit{document level} deduplication steps of FineWeb for several reasons: they may not represent a significant enough portion of a document or a single document may include repeated sections from various documents. 

We apply exact substring deduplication to remove any substring of predetermined length that repeats verbatim more than once by adapting the implementation from \cite{lee2021deduplicating} based on Suffix arrays \cite{manber1993suffix}. 
Exact substring deduplication can be fine tuned through two hyper-parameters: length-threshold (the minimum length of repeated text sequences) and frequency-threshold. We utilize a length-threshold of 50, consistent with the implementation in \cite{lee2021deduplicating, penedo2023refinedweb}. 

We make several modifications to the exact substring deduplication implementation from \cite{lee2021deduplicating} to run at scale. Furthermore, we adapt it to remove exact substring duplicates in a sharded manner. In particular, we shard each snapshot of FineWeb-V1.1.0 into sets of roughly equal size and apply exact substring deduplication on each shard independently. Also, rather than removing all copies of a duplicate substring, we retain the first occurrence of each duplicate substring and remove any subsequent matches exceeding 50 consecutive tokens.


\subsection{FastText Classifiers}
\label{sec:fastText}
FastText \cite{joulin2017bag} family of binary classifiers have been used in prior datasets \cite{weber2024redpajama, li2024datacomplm} for identifying high-quality pre-training documents. Recently, \cite{li2024datacomplm} showed that fastText classifier trained on carefully selected data can outperform sophisticated model-based filtering approaches such as AskLLM (prompting an LLM to ask if a document is helpful). Inspired by their effectiveness coupled with the computational efficiency of fastText classifiers, we use fastText classifiers for quality annotations. 

We employ two fastText classifiers: (i) the fastText classifier from \cite{li2024datacomplm} trained on a mix of instruction-formatted data (OpenHermes-2.5 \cite{openhermes2023}) and high-scoring posts from ELI5 subreddit \cite{fan2019eli5} and (ii) our own fastText classifier trained on a mix of high-quality synthetic data and data annotated by an LLM for high educational value.

Specifically, we use the supervised fastText package from \cite{joulin2017bag} to train a classifier on 400k documents, equality split between positive (i.e., high-quality) and negative (i.e., low-quality) classes, selected as follows. 
\begin{itemize}
\setlength\itemsep{0em}
\item Positive documents: 
    \begin{itemize}
    \setlength\itemsep{0em}
    \item 190k synthetic documents randomly sampled from the Cosmopedia dataset -- an open synthetic dataset consisting of textbooks, blogposts, stories, posts and WikiHow articles generated by Mixtral-8x7B-Instruct-v0.1 \cite{benallal2024cosmopedia}.
    \item 10k documents with high educational value selected as follows: we annotated 600k random documents from FineWeb-V1.1.0 asking Mixtral-8x22B-Instruct to score each document between 1 to 5  for its educational quality (with 5 being the highest quality), using a prompt similar to the one used by FineWeb-Edu. Next, we selected 10k random documents from the documents with scores $\geq 4$.
    \end{itemize}
\item Negative documents: 200k random documents out of the 600k Mixtral-annotated documents with scores $\leq 2$. 
\end{itemize}
We denote the DCLM-fastText as $\fasttexDCLM$ and our custom fastText as $\fasttexCosmo$. Each fastText classifier takes as input a document $D$ and produces a confidence score between $[0,1]$ for the document to have positive label (i.e., high-quality).\footnote{A fastText classifier conventionally outputs a label (positive or negative) along with the confidence score which can be easily converted to obtain the confidence score for the positive label.} 
In Appendix C, we present several examples showing how our custom fastText filter complements the DCLM-fastText filter.

\subsection{Readability Scores}
\label{sec:readability}
Readability scores are formulas based on text statistics (such as sentence length, average number of words, number of syllables etc.) designed to assess how easily the text can be read and understood \cite{duffy1985readability}. We apply readability scores as a novel quality metric to facilitate identifying and filtering hard-to-read low-quality documents. 

A large number of readability score formulas have been developed to asses text difficulty \cite{sarin2023can,begeny2014readability}. We experimented with a number of readability score formulas and selected McAlpine-EFLAW readability score \cite{mcalpine2006from,mueller2012mcalpine}. McAlpine-EFLAW readability score of a document is a numerical score computed as a function of the number of words in a document plus the number of mini-words (consisting of $\leq 3$ characters) divided by the number of sentences. Lower score means the document is easier to understand for a reader with English as a foreign language. Unlike other readability score formulas (such as Flesch-Kincaid \cite{kincaid1975derivation} or Gunning Fog \cite{gunning1952technique}) which are restricted to estimate a grade level for the text, McAlpine-EFLAW produces a numerical score assessing readability for a global audience \cite{sarin2023can}, making it more suitable for document quality annotation. We also demonstrate the effectiveness of the McAlpine-EFLAW score compared to other readability scores through ablation experiments. Specifically, we tested a few of readability score metrics including Flesch-Kincaid-grade level \cite{kincaid1975derivation}, Automated Readability Index (ARI)
\cite{ARI}, Gunning Fog \cite{gunning1952technique} and McAlpine-EFLAW, and determined that McAlpine-EFLAW yields the best results.


We analyzed readability score distributions of the documents grouped by categories. Specifically, we considered the documents from the following 3 snapshots from FineWeb-V1.1.0: CC-MAIN-2024-10, CC-MAIN-2023-40 and CC-MAIN-2023-14 and computed the top-level category for each document using the WatsonNLP hierarchical text categorization \cite{WatsonNLPCat}. The WatsonNLP categorization is based on the Interactive Advertising Bureau (IAB) Tech Lab categories taxonomy \cite{IABCat}. We observe the readability score distributions in certain categories, such as science, education, technology and medical health differ from the overall distribution across all categories. 
This variation in distributions can be attributed to the observation that several documents in these categories demand a higher level of education to understand and have high readability score (higher the readability score, more difficult is the English document to read), leading to a higher average readability score. 

Based on this observation, there is a risk of losing high-quality documents if a threshold is selected based on the overall data distribution and the same threshold is applied to all documents. Guided by readability score distributions in different categories, we leverage the category information of documents and develop a category-aware readability score quality filter as part of our ensemble quality filter (Section~\ref{sec:ensemble_filter}). In general, we use a more lenient threshold for these specific categories to prevent filtering out documents with potential educational value solely because of their high readability scores which results in better performance compared to filtering without leveraging category information.  We also performed ablations with other categories. For example, adding ``news and politics", ``business and finance" as well as ``personal finance" to the hard to read categories degraded performance (Section~\ref{sec:ablation_readability_scores}).
In Appendix C, we present several low quality examples detected and filtered out by our category-aware readability score filter.



\subsection{Extreme-Tokenized Documents}
\label{sec:extreme_tokenized}
After manually inspecting fastText model-quality annotations and readability scores of large number of low-quality documents, we found that several abnormal documents were mislabeled by these annotators. We observed a peculiar pattern after tokenizing these documents: while most of these documents had similar lengths, they produced significantly different token counts. To quantify this effect, we propose novel annotations that effectively leverages information from the ``pre-tokenization'' stage (document char length, document size) and the ``post-tokenization'' stage (token counts) to identify potential low-quality documents. 

\begin{figure*}
\centering
  \includegraphics[width=0.8 \textwidth]{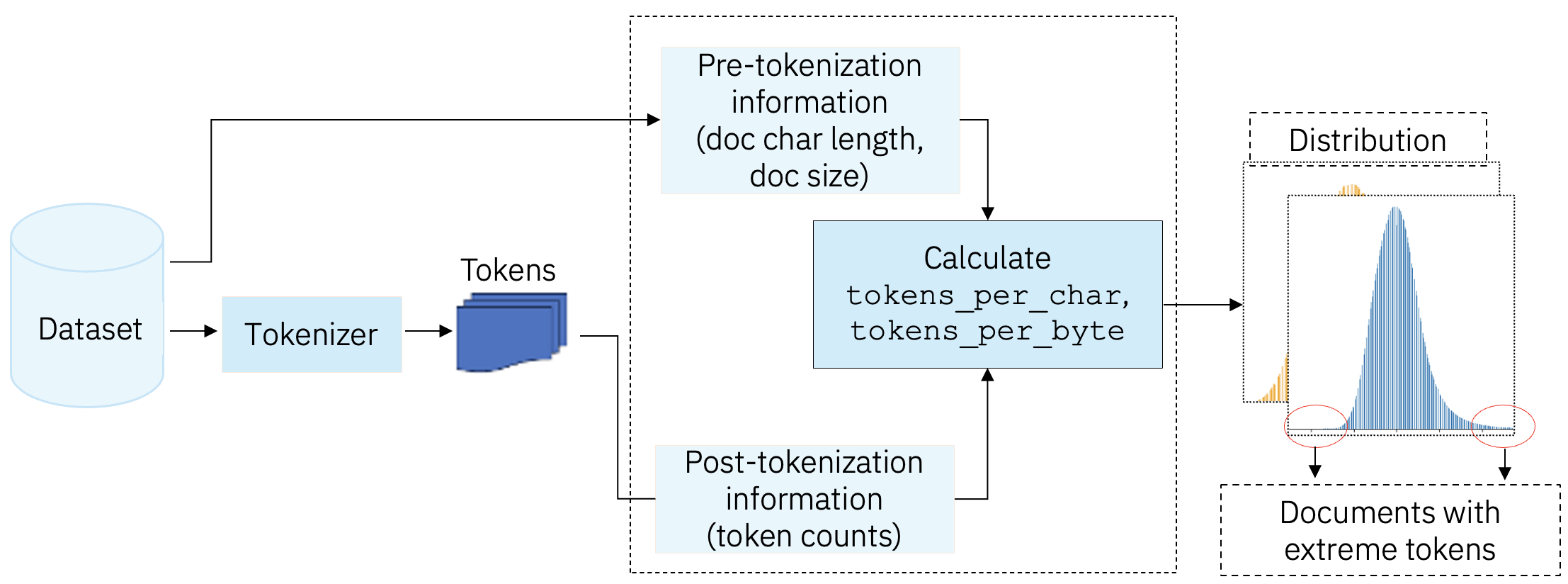}
  \caption{Sequence of steps for removing extreme tokenized documents.}
  \label{fig:extreme_tokenized_schematic}
\end{figure*}

Specifically, for each document $D$, we compute the the following two annotations:
\begin{align*}
    \text{TokensPerChar}(D) = \frac{\text{Number of Tokens in $D$}}{\text{Number of Characters in $D$}}, \:
     \text{TokensPerByte}(D) = \frac{\text{Number of Tokens in $D$}}{\text{Size of $D$ (in bytes)}}.
\end{align*}
We refer to the the documents with extremely high or low number of tokens per character (or tokens per byte) as \textit{extreme-tokenized} documents (see Fig.~\ref{fig:extreme_tokenized_schematic} for a schematic).  

Data quality filtering based on tokenized data has been used in other works \cite{mehta2024openelm, soldaini2024dolma} to improve the data quality by filtering out documents with too few tokens \cite{soldaini2024dolma} or removing the sequences containing fewer tokens than a specified threshold. However, the effectiveness of these approaches in detecting low-quality documents is limited because of their sole reliance on the token count. Our extreme-tokenized quality filter does not solely rely on token count but also effectively leverages both information from the ``pre-tokenization'' stage and the ``post-tokenization'' stage to identify and filter out low-quality documents.

We analyzed the distributions of TokensPerChar and TokensPerByte for documents grouped by category. Specifically, we considered the documents from the following 3 snapshots from FineWeb-V1.1.0: CC-MAIN-2024-10, CC-MAIN-2023-40 and CC-MAIN-2023-14, and computed the top-level category for each document using the WatsonNLP hierarchical text categorization \cite{WatsonNLPCat}, which is based on the Interactive Advertising Bureau (IAB) Tech Lab categories taxonomy \cite{IABCat}. We observe that the distributions are generally bell-shaped for each category, but the values of the mean and variance differ by category. Furthermore, we observe that low-quality documents typically fall into the two extremes of the distribution. Therefore, we characterize extreme-tokenized documents of a given category as those falling into the two extremes of the TokensPerChar (or TokensPerByte) distribution for the category.
Guided by the distributions of TokensPerChar and TokensPerByte in different categories, we leverage the category information of documents and develop a category-aware extreme-tokenized quality filter as part of our ensemble quality filter (Section~\ref{sec:ensemble_filter}). At a high level, we use stricter thresholds on TokensPerChar/TokensPerByte for documents outside the key categories and use more lenient thresholds for documents in these key categories (Section~\ref{sec:ablation_extreme_tokenized}).
In Appendix C, we present several low quality examples detected and filtered out by our category-aware Extreme-Tokenized documents filter.
 



\subsection{Document Category Classifiers}
\label{sec:category_classifiers}
As mentioned in previous sections, the quality score distributions of documents in certain categories, which tend to contain documents with high educational-level, differ from the overall distribution across all categories. In particular, we observe that the following IAB categories supported by WatsonNLP categorization have significantly different distributions than the overall distribution across all categories: science, education, technology \& computing, and medical health. Thus, for each of these key categories, we annotate whether each document falls into the category. 


To perform category classification on the 96 snapshots in FineWeb-V1.1.0 at scale, we train four binary fastText category classifiers for each of the four key categories. Specifically, we generated labeled data using the WatsonNLP hierarchical categorization \cite{WatsonNLPCat}, and used the supervised fastText package from \cite{joulin2017bag} to train the fastText classifiers on the following documents:
\begin{itemize}
\setlength\itemsep{0em}
\item Positive documents: 400k documents randomly sampled from the documents labeled with that specific category with a confidence score 0.95 and above.
\item Negative documents: 400k documents randomly sampled from the documents labeled with any category other than these four categories with a confidence score of 0.95 and above.
\end{itemize}
We denote the fastText classifiers as $\sciClassifier$, $\eduClassifier$, $\techClassifier$, and $\medClassifier$. Each classifier takes as input a document and produces a label whether the document belongs to the category, along with a confidence score between $[0,1]$.

We use our trained document category classifiers to annotate all the snapshots from FineWeb-V1.1.0.
We leverage these category annotations in our category-aware readability score quality filtering and extreme-tokenized quality filtering which results in better performance compared to filtering without leveraging category information.

\subsection{Ensemble Quality Filter}
\label{sec:ensemble_filter}
Equipped with multiple quality annotators, we develop an ensemble quality filter with the aim of maximizing data quality under the constraint of retaining nearly 10T tokens from FineWeb-V1.1.0. We construct our ensemble quality filter by selecting thresholds for individual annotators and then designing an ensemble filtering rule for aggregating the filter outputs.

Specifically, we select the thresholds on readability scores integrating the category annotations to design Category-Aware Readability Score filter. We choose our initial thresholds based on the readability score distributions for key categories (computed on entire FineWeb-V1.1.0), and subsequently fine-tune them through ablation experiments to identify the best set of thresholds that result in maximum performance gain (see Section~\ref{sec:ablation_readability_scores}). Similarly, we select the thresholds for Category-Aware Extreme-Tokenized Documents filter (see Section~\ref{sec:ablation_extreme_tokenized}). Then, given an aggregation rule, we choose the thresholds for fastText filters such that we retain nearly 10T tokes from FineWeb-V1.1.0. As an example, a simple aggregation rule is to apply each filter sequentially (which essentially is a logical AND of filter outputs).

\begin{figure}[!h]
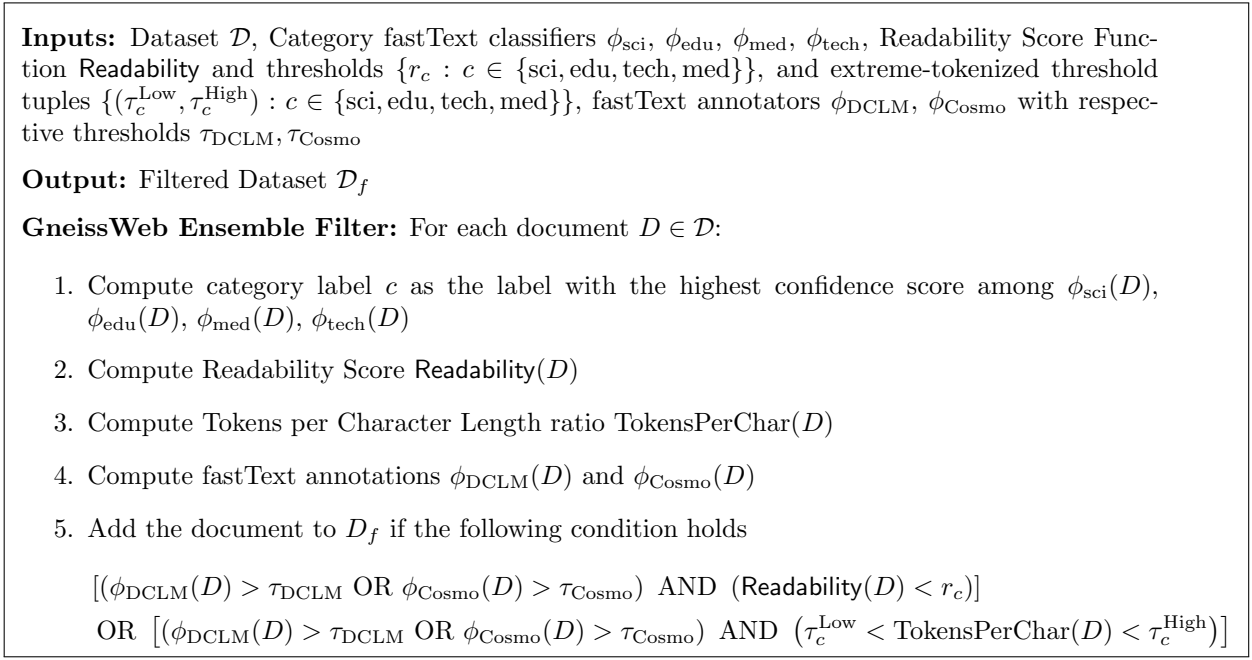

\fbox{
	\begin{minipage}{0.97\textwidth}
	\myparagraph{Inputs:} Dataset $\mathcal{D}$, Category fastText classifiers $\sciClassifier$, $\eduClassifier$, $\medClassifier$, $\techClassifier$, Readability Score Function $\readability$ and thresholds $\{\thresholdRscoreCat : c\in\{\textrm{sci},\textrm{edu},\textrm{tech},\textrm{med}\}\}$,  and extreme-tokenized threshold tuples $\{(\thresholdETCatLow, \thresholdETCatHigh) : c\in\{\textrm{sci},\textrm{edu},\textrm{tech},\textrm{med}\}\}$, fastText annotators $\fasttexDCLM$, $\fasttexCosmo$ with respective thresholds $\thresholdDCLM, \thresholdCosmo$
	
    \myparagraph{Output:} Filtered Dataset $\mathcal{D}_f$

    \myparagraph{GneissWeb Ensemble Filter:} For each document  $D\in\mathcal{D}$:
    	
	\begin{enumerate}
	 \item Compute category label $c$ as the label with the highest confidence score among $\sciClassifier(D)$, $\eduClassifier(D)$, $\medClassifier(D)$, $\techClassifier(D)$
     \item Compute Readability Score $\readability(D)$
     \item Compute Tokens per Character Length ratio $ \textrm{TokensPerChar}(D)$
     \item Compute fastText annotations $\fasttexDCLM(D)$ and $\fasttexCosmo(D)$
     \item Add the document to $D_f$ if the following condition holds
     \begin{align*}
         & \left[\left(\fasttexDCLM(D) > \thresholdDCLM \textrm{ OR } \fasttexCosmo(D) > \thresholdCosmo\right) \textrm{ AND } \left(\readability(D) < \thresholdRscoreCat\right)\right]\\ 
         & \textrm{ OR }  \left[\left(\fasttexDCLM(D) > \thresholdDCLM \textrm{ OR } \fasttexCosmo(D) > \thresholdCosmo\right) \textrm{ AND } \left(\thresholdETCatLow < \textrm{TokensPerChar}(D) < \thresholdETCatHigh  \right)\right]
     \end{align*}
	\end{enumerate}
 	\end{minipage}
 	}
\caption{
	GneissWeb Ensemble Quality Filter}
	\label{fig:gneissweb_ensemble}
\end{figure}

We perform ablations on a variety of aggregation rules and determine the \textit{best} aggregation rule that provides the maximum performance gain. We provide the details of our ensemble quality filter in Fig.~\ref{fig:gneissweb_ensemble}. For the category-aware extreme-tokenized documents filter, we only used TokensPerChar heuristic for our final recipe, as both TokensPerByte and TokensPerChar showed similar distributions.

We provide in detail various ablation experiments in evaluating the impact of our ensemble based filtering rule in Sec.~\ref{sec:ablations} and provide the comparisons with other combinations of the key components in Appendix.


\subsection{Putting It All Together}
\label{sec:gneissweb_recipe}

The GneissWeb recipe consists of first applying the exact substring deduplication, computing category and quality annotations, and then applying the ensemble quality filter as shown in Fig.~\ref{fig:GneissWeb}. We obtain the GneissWeb dataset of 10T tokens by applying the GneissWeb recipe to the 15T tokens in the 96 snapshots of FineWeb-V1.1.0. We prepared GneissWeb using a version of IBM's DataPrep kit library \cite{wood2024data} which will be released in open source in future.

We note that, while the GneissWeb recipe is designed with the goal of obtaining $\sim$10T high quality tokens suitable for Stage-1 pre-training, it is also possible to adapt the recipe by tuning filtering parameters to produce smaller and higher quality datasets fit for Stage-2 type of pre-training.

\section{Experiments}
\label{sec:experiments}

\subsection{Ablation and Evaluation Setup}
\label{sec:setup}
We analyze our recipe ingredients and design choices by training data ablation models that are identical in terms of architecture and training parameters, except for the data they were trained on. We evaluate the ablation models on a wide range of downstream benchmarks (details below). 
\\

\noindent\textbf{Training:} To minimize the impact of  random data subset selection on evaluation scores, we use three equal-sized random subsets of the full data to train three models, and compute average scores along with standard deviation.
More specifically, when comparing two dataset versions $\mathcal{D}_1$ and $\mathcal{D}_2$, we select three equal-sized random subsets $D^i_1, D^i_2, D^i_3$ from each $\mathcal{D}_i, i\in\{1,2\}$, and train three models using the random subsets. We compare the average scores across the three models and also report standard deviations. 

Following prior ablations in open datasets \cite{penedo2023refinedweb,penedo2024fineweb,li2024datacomplm}, we train decoder-only models with Llama architecture \cite{touvron2023llama2}. 
We adopt 1.4B parameter models (including embeddings) for the majority of our experiments and perform training with a sequence length of 8192, a global batch size of $\sim 1$ million tokens, and the StarCoder tokenizer \cite{li2023starcoder}.
In our ablation experiments, we typically train the models on 35B (slightly larger than the Chinchilla optimal) tokens, similar to \cite{penedo2023refinedweb,penedo2024fineweb}. 
In our main experiments comparing our dataset with other open-source datasets, we train the models on 350B tokens, similar to \cite{penedo2024fineweb}. In addition, to evaluate our dataset for training larger models, we perform controlled ablations by training models with 3B and 7B parameters on 100B tokens.\\

\noindent\textbf{Evaluation:} 
We evaluate our models using LM Evaluation Harness \cite{gao2024lmeval} on two categories of tasks: 11 \textit{High-Signal tasks} (18 variants combining 0-shot and few-shot) and 20 \textit{Extended tasks} (29 variants combining 0-shot and few-shot). Throughout the training, we evaluate intermediate model checkpoints on high-signal tasks, and evaluate the final checkpoints on high-signal as well as extended tasks. For more details on the tasks, see Appendix~\ref{app:evaluation_benchmarks}.

\textit{High-Signal tasks:} Since ablations are performed by training `small' models (1.4B parameter models) for a `few billion' tokens (typically 35B tokens), it is important to identify benchmarks that provide good signal at this relatively small scale. Similar to \cite{penedo2024fineweb}, we use the criteria of accuracy above random guessing, accuracy increases over training, and small variance across runs to select 11 High-Signal/Early-Signal tasks. We use both the zero-shot as well as few-shot variations of these tasks for 18 variants in total (Appendix~\ref{app:evaluation_benchmarks}).  

\textit{Extended tasks:} We evaluate the final checkpoints of our models on 20 tasks with 29 variants combining zero-shot and few shot. This broader set of tasks are useful indicators for larger model performance and thus have retained in the Extended Tasks set (Appendix~\ref{app:evaluation_benchmarks}).

\begin{figure*}[t!]
    \centering
    \begin{subfigure}[t]{0.5\textwidth}
        \centering
        \includegraphics[scale=0.525]{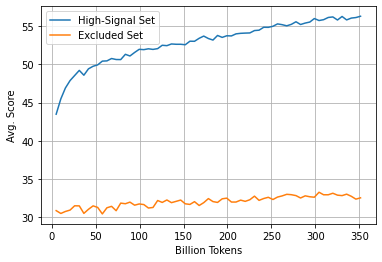}
    \end{subfigure}%
    ~ 
    \begin{subfigure}[t]{0.5\textwidth}
        \centering
        \includegraphics[scale=0.525]{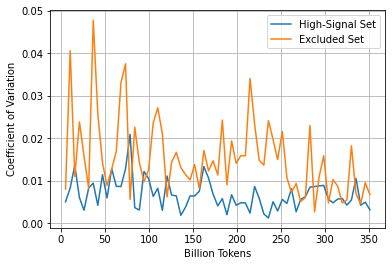}
    \end{subfigure}
    \caption{High signal tasks provide early performance indication for small models at few billion tokens. They also show smaller variation in performance for models trained on random subsets. See Appendix~\ref{app:evaluation_benchmarks} for the full list of tasks.}
    \label{fig:hs_vs_excluded_tasks}
\end{figure*}

These differences between the High-Signal Tasks vs Extended Tasks are seen in Fig.~\ref{fig:hs_vs_excluded_tasks}, where we see a comparison of the High Signal Tasks versus those which are in the Extended Tasks and excluded from the High Signal Tasks.  We observe that the average accuracy increases in the former and is relatively static in the latter. This was a criteria for excluding them from the High Signal Task set.

The high signal tasks also show lower coefficient of variation compared to the excluded tasks as shown in Fig.~\ref{fig:hs_vs_excluded_tasks}. The coefficient of variation is calculated as the ratio between the standard deviation of the average score divided by the mean, where statistics are computed across models trained on three random subsets of equal size. Lower coefficient of variation shows more stable results, due to lower variance across random subsets. Their lower coefficient of variation makes the high-signal tasks more reliable at the ablation scale.

\subsection{Compute Infrastructure}

We train and evaluate our models on an LSF (Load Sharing Facility) cluster comprising multiple Dell XE9680 nodes, each equipped with eight H100 GPUs. For training tasks involving 35 billion tokens, we typically use models with 1.4 billion trainable parameters across 64 GPUs (or 8 nodes). For more intensive tasks, we scale up to 128 or 256 GPUs to reduce training time. Evaluation tasks are primarily run on a single node with 8 GPUs. 

The entire model training and evaluation process is fully automated using GitOps, with progress updates frequently sent to a Slack channel. A user begins by specifying model configurations and datasets in a \texttt{.yaml} file and submitting it for review via a pull request to a GitHub repository. Once approved, the system automatically submits the job if the requested resources are available. For datasets stored in COS (Cloud Object Storage), the system first downloads them to IBM's GPFS (General Parallel File System) to minimize network traffic during training. With this computational infrastructure, the training speed of an FSDP model with 1.4 billion parameters is approximately 32,000 tokens per GPU per second. Consequently, training the model with 35 billion tokens typically takes about 4.6 hours when utilizing 64 GPUs. Model checkpoints are saved at regular intervals (based on the number of trained tokens) and evaluated in real time, with the results automatically pushed to IBM's lakehouse for querying and visualization. Throughout each stage, the user receives updates in the Slack channel, ensuring transparency and progress tracking throughout the process.

\subsection{Evaluating the GneissWeb Dataset}
\label{sec:evaluation}

We compare our GneissWeb dataset with the following state-of-the-art open-source, web-scale datasets: FineWeb (15T tokens) \cite{penedo2024fineweb}\footnote{We used FineWeb-V1.1.0 \url{https://huggingface.co/datasets/HuggingFaceFW/fineweb}}, FineWeb-Edu-Score-2 (5.4T tokens) \cite{penedo2024fineweb}, DCLM-Baseline (3.8T tokens) \cite{li2024datacomplm}, Dolma (3T tokens), FineWeb-Edu (1.3T tokens) \cite{penedo2024fineweb}, and RefinedWeb (600B tokens) \cite{penedo2023refinedweb}.  

\subsubsection{1.4B Models Trained on 350B Tokens}
\label{sec:1.4B_models_350B_tokens}

Table~\ref{table:1.4B} shows the average scores on high-signal tasks and extended tasks for 1.4 billion parameter models trained on three randomly sampled sets of 350B tokens from each dataset. The datasets evaluated are broken down into those which are above 5 trillion tokens (highlighted in blue) in size and those below 5 trillion. The former are useful for Stage-1 kind of training and are the primary focus of this study. The latter are useful for Stage-2 kind of training and with certain tuning of parameters of filtering a version of GneissWeb can be produced for this space. GneissWeb demonstrates the best performance among large datasets. Specifically, models trained on the GneissWeb outperform those trained on FineWeb-V1.1.0 by 2.14 percent points on high-signal tasks, and by 1.49 percent points on extended tasks. 

For datasets that are greater than 5 trillion token set size, in Table \ref{table:AvgCat}, we show the performance broken down into the various categories of tasks -- Commonsense Reasoning (CR),  Language Understanding (LU), Reading Comprehension (RC), World Knowledge (WK) and Symbolic Problem Solving (SPS).  As shown in Table \ref{table:AvgCat},  GneissWeb is not only the best overall but in fact performs the best in all categories of tasks except World Knowledge.

\begin{figure*}[!t]
  \centering 
  \includegraphics[width=0.7 \textwidth]{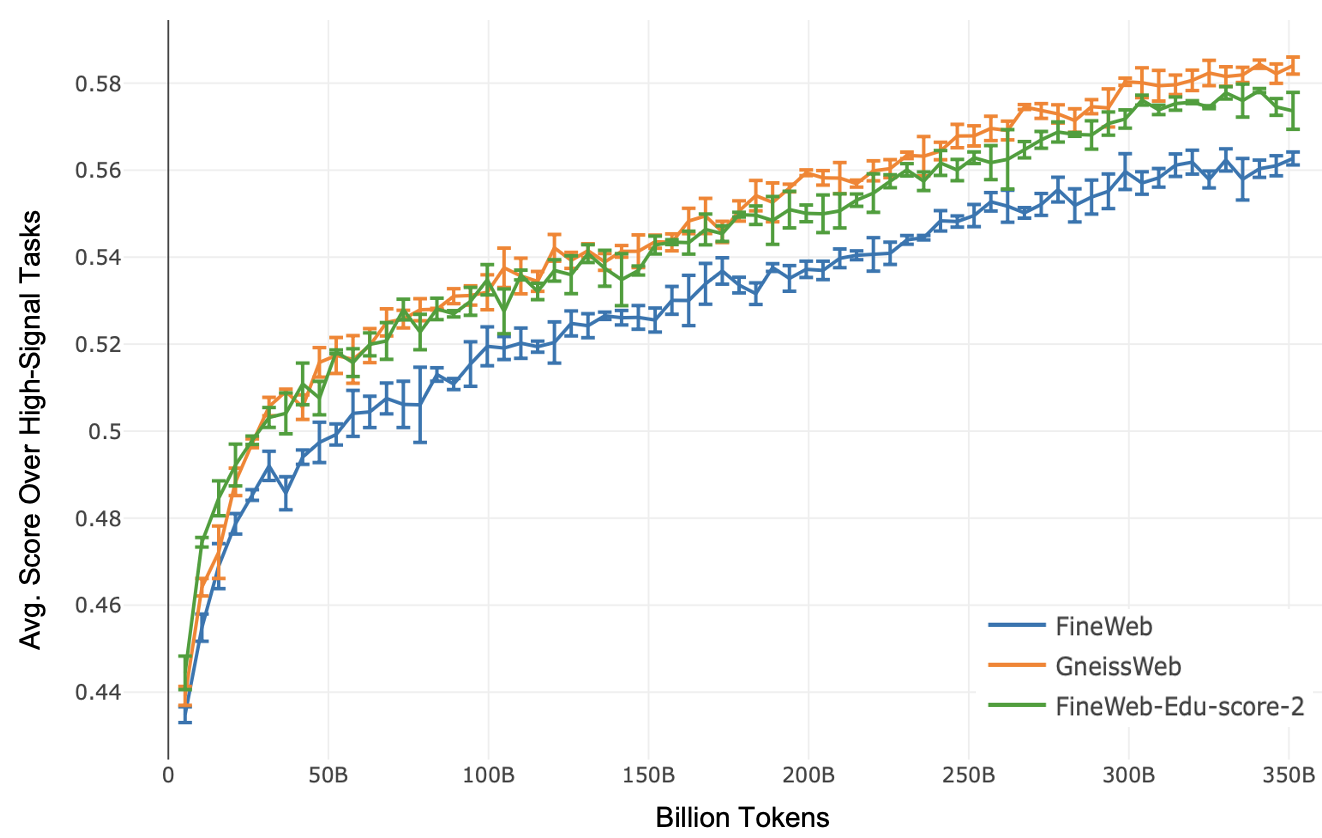}
  \caption{Average evaluation score on High-Signal tasks versus the number of tokens for 1.4 Billion parameter models. The models trained on GneissWeb consistently outperform the ones trained on FineWeb.V1.1.0 and FineWeb-Edu-score-2.}
  \label{fig:1-4B}
\end{figure*}

In Figure \ref{fig:1-4B}, we show the progression of average score over high-signal tasks with training for 1.4~billion parameter model for 350 billion tokens. We see that for all three datasets compared, the accuracy increases over time and the accuracy of GneissWeb is consistently higher than FineWeb.V1.1.0 and FineWeb-Edu-score-2.

\begin{table*}
\centering
\begin{tabular}{c c c c}
\toprule
  Dataset & Tokens & High-Signal Eval Score & Extended Eval Score \\
\midrule
\rowcolor{blue!15}
FineWeb-V1.1.0 & 15T & 56.26 $\pm$ 0.14 & 47.33 $\pm$ 0.3 \\
\rowcolor{blue!15}
\textbf{GneissWeb} & \textbf{9.8T} & \textbf{58.40 $\pm$ 0.19} & \textbf{48.82 $\pm$ 0.27} \\
\rowcolor{blue!15}
FineWeb-Edu-Score-2 & 5.4T & 57.36 $\pm$ 0.42 & 48.16 $\pm$ 0.29 \\
DCLM-Baseline & 3.8T & 61.36 $\pm$ 0.11 & 51.09 $\pm$ 0.42 \\
Dolma & 3T & 54.18 $\pm$ 0.65 & 47.39 $\pm$ 0.75 \\
FineWeb-Edu & 1.3T & 58.44 $\pm$ 0.14 & 48.91 $\pm$ 0.13 \\
RefinedWeb & 0.6T & 57.77 $\pm$ 0.10 & 48.11 $\pm$ 0.3 \\
\bottomrule
\end{tabular}
\caption{\textbf{Comparison of the GneissWeb dataset with other public datasets.} Average scores of 1.4 Billion parameter models trained on 350 Billion tokens randomly sampled from state-of-the-art open datasets. Scores are averaged over 3 random seeds used for data sampling and are reported along with standard deviations.  GneissWeb performs the best among the class of large datasets.}
\label{table:1.4B}
\end{table*}

\begin{table*}[t]
\small
\centering
\begin{tabular}{c c c c c c c}
\toprule
{Dataset} & \begin{tabular}{@{}c@{}}Commonsense \\ Reasoning\end{tabular} & \begin{tabular}{@{}c@{}}Language\\ Understanding\end{tabular} & \begin{tabular}{@{}c@{}}Reading\\ Comprehension\end{tabular} & \begin{tabular}{@{}c@{}}World\\ Knowledge\end{tabular} & \begin{tabular}{@{}c@{}}Symbolic\\ Problem\\ Solving\end{tabular} & {Average} \\
\midrule
FineWeb.V1.1.0  & 45.23 & 47.58 & 62.67 & 39.01 & 26.16 & 47.17 \\
\textbf{GneissWeb} & \textbf{45.53} & \textbf{48.77} & \textbf{65.21} & \textbf{41.09} & \textbf{27.92} & \textbf{48.82} \\
FineWeb-Edu-Score-2 & 45.32 & 47.2 & 63.29 & 42.24 & 27.25 & 48.16 \\
\bottomrule
\end{tabular}
\caption{\textbf{GneissWeb outperforms other large public datasets (5T+ tokens) on most categories.} Average evaluation scores grouped by categories for 1.4 Billion parameter models trained on 350 Billion tokens (see Appendix~\ref{app:evaluation_benchmarks} for the tasks in each category).}
\label{table:AvgCat}
\end{table*}


\begin{table*}[t]
\centering
\begin{tabular}{c c c c}
\toprule
  Dataset & High-Signal Eval Score & Extended Eval Score \\
\midrule
FineWeb.V1.1.0 & 60.31 $\pm$ 0.21 & 50.15 $\pm$ 0.07 \\
\textbf{GneissWeb} & \textbf{62.83 $\pm$ 0.24} & \textbf{52.1$\pm$0.22} \\
FineWeb-Edu-Score-2 & 61.63 $\pm$ 0.04 & 51.13 $\pm$ 0.17 \\
\bottomrule
\end{tabular}
\caption{\textbf{GneissWeb outperforms other large public datasets (5T+ tokens) at 3B model size.} Average Eval Scores on High Signal and Extended Tasks for 3B models trained on 350B tokens.  Scores are averaged over 3 random seeds used for data sampling and are reported along with standard deviations.}
\label{table:3B}
\end{table*}


\begin{table*}
\centering
\begin{tabular}{c c c c}
\toprule
  Dataset & High-Signal Eval Score & Extended Eval Score \\
\midrule
FineWeb.V1.1.0 & 64.61 $\pm$ 0.23 & 53.39 $\pm$ 0.25 \\
\textbf{GneissWeb} & \textbf{67.34 $\pm$ 0.26} & \textbf{55.14 $\pm$ 0.28} \\
FineWeb-Edu-Score-2 & 65.51 $\pm$ 0.34 & 54.61 $\pm$ 0.31 \\
\bottomrule
\end{tabular}
\caption{\textbf{GneissWeb outperforms other large public datasets (5T$+$ tokens) at 7B model size.} Average  Scores on High Signal and Extended Tasks for 7B models trained on 350B tokens.  Scores are averaged over 3 random seeds used for data sampling and are reported along with standard deviations.}
\label{table:7B}
\end{table*}


\subsubsection{3B and 7B Models Trained on 350B Tokens}
\label{sec:3b_7b_model_100b_tokens}
To evaluate the GneissWeb for training larger models, we perform controlled ablations by training models with 3 billion and 7 billion parameters on 350 billion tokens.
Given that training models of size 3 and 7 billion parameters require lot more compute and so does evaluation, we have 
restricted comparison with large datasets (FineWeb and FineWeb-Edu-Score-2). Specifically, we train models on three randomly sampled sets of 350 billion tokens from each dataset and compute the average scores.

\begin{figure*}[!t]
\begin{minipage}[t]{0.45\linewidth}
  \includegraphics[width=\textwidth]{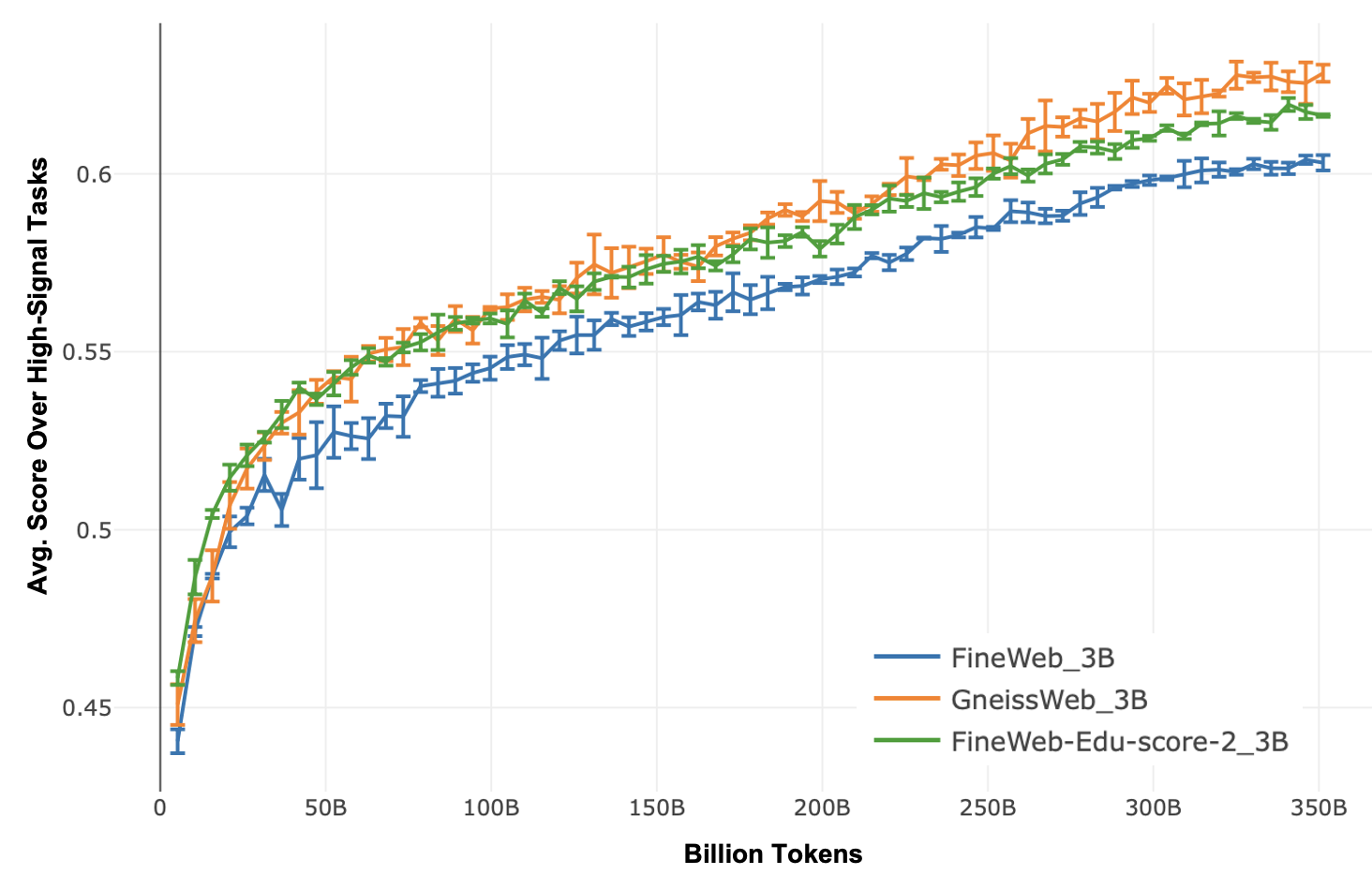}
  \caption{Average evaluation score on High-Signal tasks versus the number of tokens at 3 Billion model size for 350 Billion tokens. The model trained on GneissWeb consistently outperforms the one trained on FineWeb.V1.1.0 throughout the training.}
  \label{fig:3B}
\end{minipage}%
    \hfill%
\begin{minipage}[t]{0.45\linewidth}
  \includegraphics[width=\textwidth]{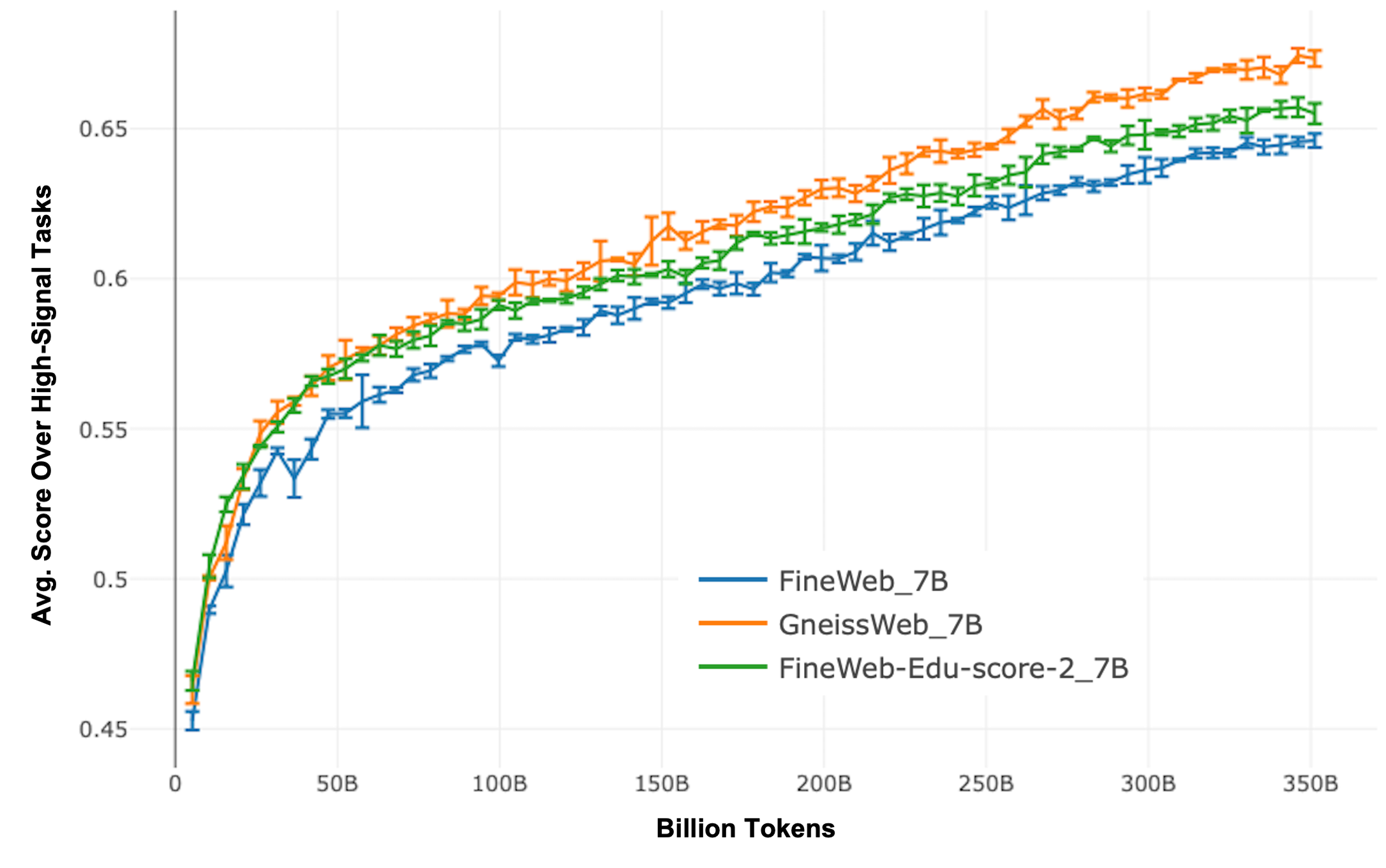}
  \caption{Average evaluation score on High-Signal tasks versus the number of tokens at 7 Billion model size for 350 Billion tokens. The model trained on GneissWeb consistently outperforms the one trained on FineWeb.V1.1.0 throughout the training.}
  \label{fig:7B}
\end{minipage}%
\end{figure*}

Table~\ref{table:3B} and Fig.~\ref{fig:3B} depict the results for 3B model size. We observe that models trained on GneissWeb outperform those trained on FineWeb.V1.1.0 by 2.52 percent points in terms of the average score computed on high-signal benchmarks (both zero-shot and few-shot), and 1.95 percent points on Extended benchmarks (both zero-shot and few-shot). 

Table \ref{table:7B} and Fig. \ref{fig:7B} show the results for 7B model size.
Models trained on GneissWeb outperform those trained on FineWeb.V1.1 by 2.73 percent points in terms of the average score computed on a set of 11 High-signal benchmarks (both zero-shot and few-shot), and 1.75 percent points on Extended benchmarks (both zero-shot and few-shot).

\subsection{Ablation Experiments}
\label{sec:ablations}
In this section, we present ablation experiments for individual ingredients as well as ensemble quality filtering. For ablations evaluating individual ingredients, we evaluate the models on a subset of 8 high-signal tasks to save compute (see Appendix~\ref{app:evaluation_benchmarks}). 

\subsubsection{Exact Substring Deduplication}
\label{sec:ablation_exact_substring_dedup}
We conduct an ablation experiment to evaluate the impact of exact substring deduplication on the model performance. As discussed in \cite{penedo2024fineweb}, the impact of deduplication is not typically visible for small number of tokens. Thus, we train two 1.4B models each on 350B tokens as follows. 
The baseline model is trained on 350B tokens randomly sampled from FineWeb-V1.1.0, and the second model is trained on the 350B tokens randomly sampled after applying sharded exact substring deduplication to FineWeb-V1.1.0 as discussed in Sec.~\ref{sec:substring_dedup}.

\begin{figure*}
  \centering 
  \includegraphics[width=0.65 \textwidth]{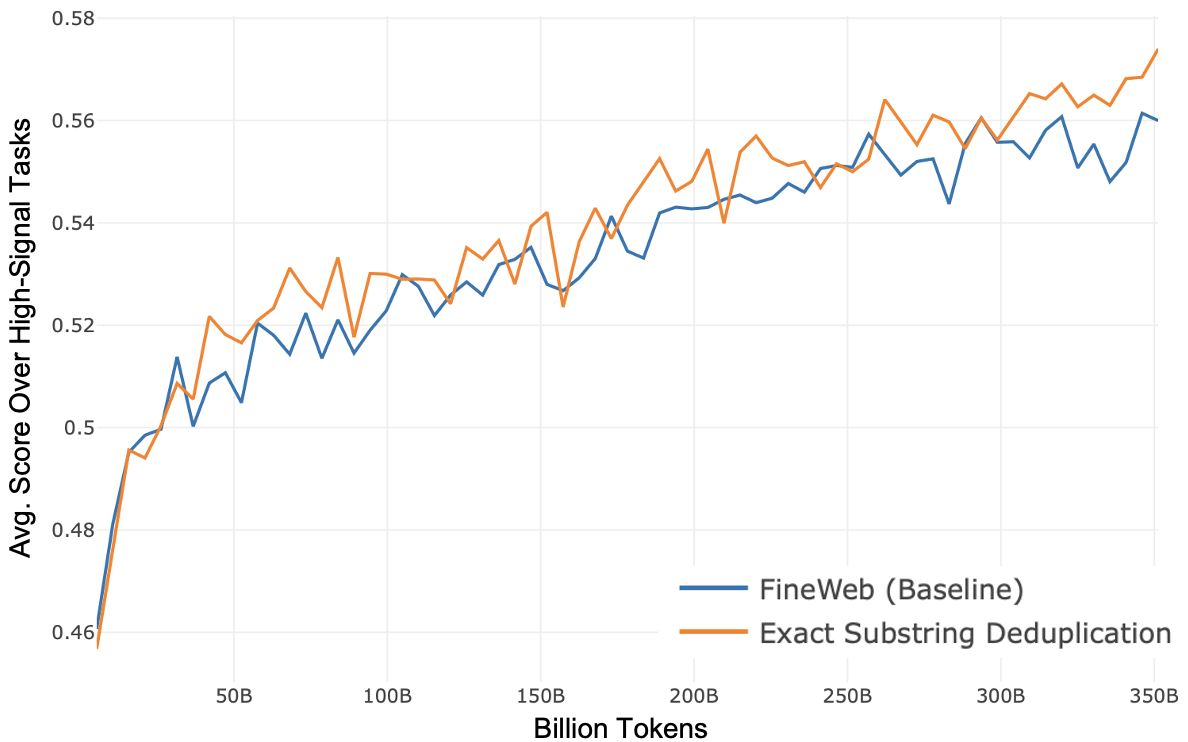}
  \caption{Ablation experiment comparing Exact Substring Deduplication against the FineWeb.V1.1 baseline at 1.4 Billion model size for 350 Billion tokens.}
  \label{fig:dedup}
\end{figure*}

In Fig.~\ref{fig:dedup}, we compare average evaluation score on high-signal tasks for the two models.  We see that for both datasets compared, the average score increases as the training progresses, and the score of the model trained on the dataset with exact substring deduplication is consistently higher (especially after 260B tokens) ending at 57.39 percent than the baseline which ends at 55.99 percent.

\subsubsection{Category-Aware Readability Score Filter}
\label{sec:ablation_readability_scores}
As discussed in Section \ref{sec:readability}, our analysis of readability score distributions of documents grouped by categories depicts that distributions of  certain categories differ from the overall distribution across categories. These specific categories tend to contain many documents with educational-style content, resulting in higher values of readability scores. Equipped with this observation, we design category-aware readability score filter wherein we select lenient filtering threshold on readability scores for documents from these educational-style categories, and stricter filtering threshold for documents outside of these categories. 
We select initial thresholds based on readability score distributions, and then perform ablations to tune the thresholds.
We use lenient threshold for the following educational-style categories: science, education, technology and computing, and medical health.
We also performed ablations to include other categories, for instance, adding ``news and politics", ``business and finance" as well as ``personal finance" to the hard-to-read categories, but it degraded performance.

In Figure \ref{fig:Rscore}, we plot the average score over high-signal tasks for the best thresholds.  Specifically,
we train two 1.4B parameter models -- the baseline model is trained on 35B tokens randomly sampled from FineWeb-V1.1.0, and the second model is trained on the 35B tokens randomly sampled after applying category-aware readability score filter to FineWeb-V1.1.0. 
We see that for both datasets compared, the average accuracy increases with training and the accuracy of the dataset with readability score quality filter is consistently higher than the baseline, achieving the final score of 53.20 percent as compare to the score of 51.94 percent for the baseline.

\subsubsection{Category-Aware Extreme-Tokenized Filter}
\label{sec:ablation_extreme_tokenized}
As mentioned in Section \ref{sec:extreme_tokenized}, we annotate each document with two parameters defined using pre-tokenization and post-tokenization document properties: TokensPerChar (number of tokens divided by number of characters) and TokensPerByte (number of tokens divided by the document size in bytes). When we plot distributions of TokensPerChar and TokensPerByte, we observe that distributions of the documents in specific education-style categories differ than the overall distribution across categories. Guided by this observation, we design our category-aware extreme-tokenized documents filter, in which, we select lenient thresholds on TokensPerChar/TokensPerByte for the specific categories and stricter thresholds for the other categories. Specifically, we select lenient thresholds for the same categories as in the case of readability scores: science, education, technology and computing, and medical health. Our ablations show that adding other categories (where distributions differ) such as personal finance degrade performance. We choose initial thresholds based on the TokensPerChar/TokensPerByte distributions, and then perform ablations to tune the thresholds. 

\begin{figure*}
\begin{minipage}[t]{0.45\linewidth}
  \includegraphics[width=\textwidth]{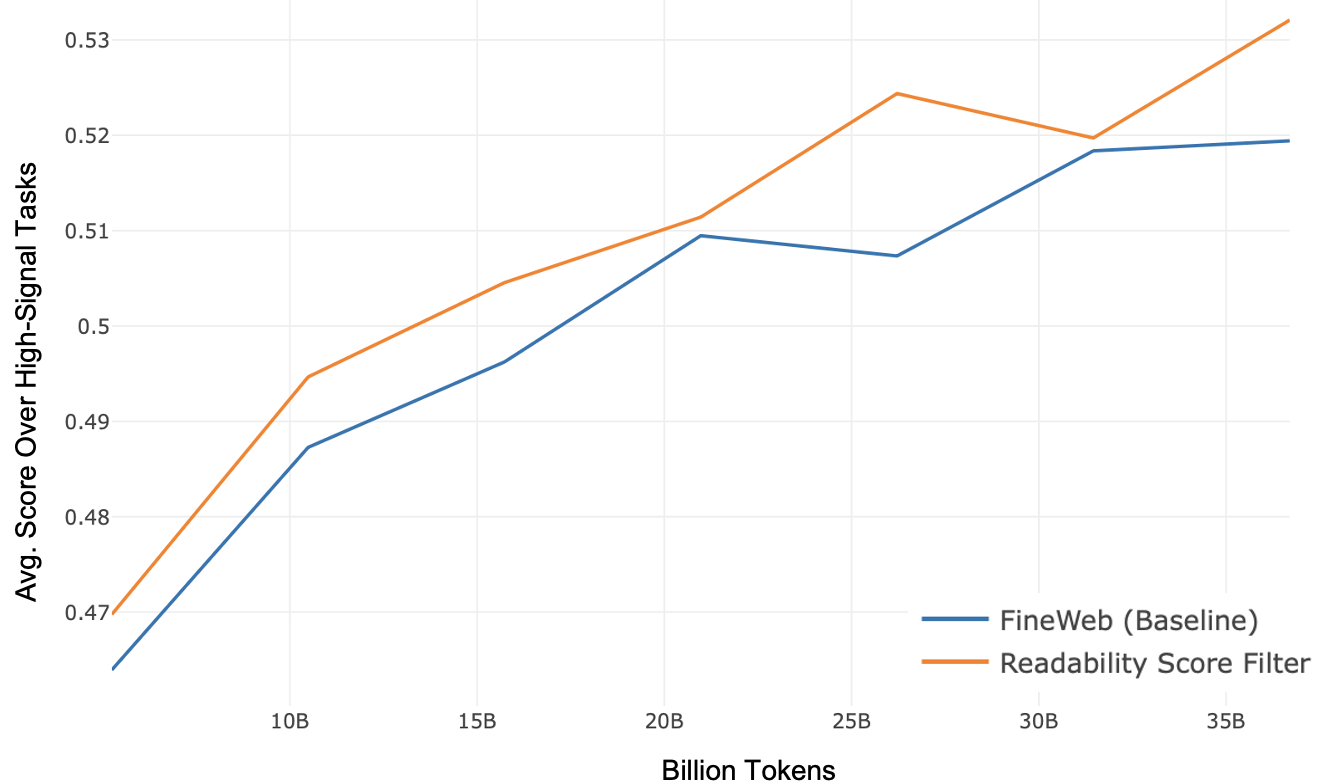}
  \caption{Ablation experiment comparing Readability Score Filter against the FineWeb.V1.1 baseline at 1.4 Billion model size for 35 Billion tokens.}
  \label{fig:Rscore}
\end{minipage}%
    \hfill%
\begin{minipage}[t]{0.45\linewidth}
  \includegraphics[width=\textwidth]{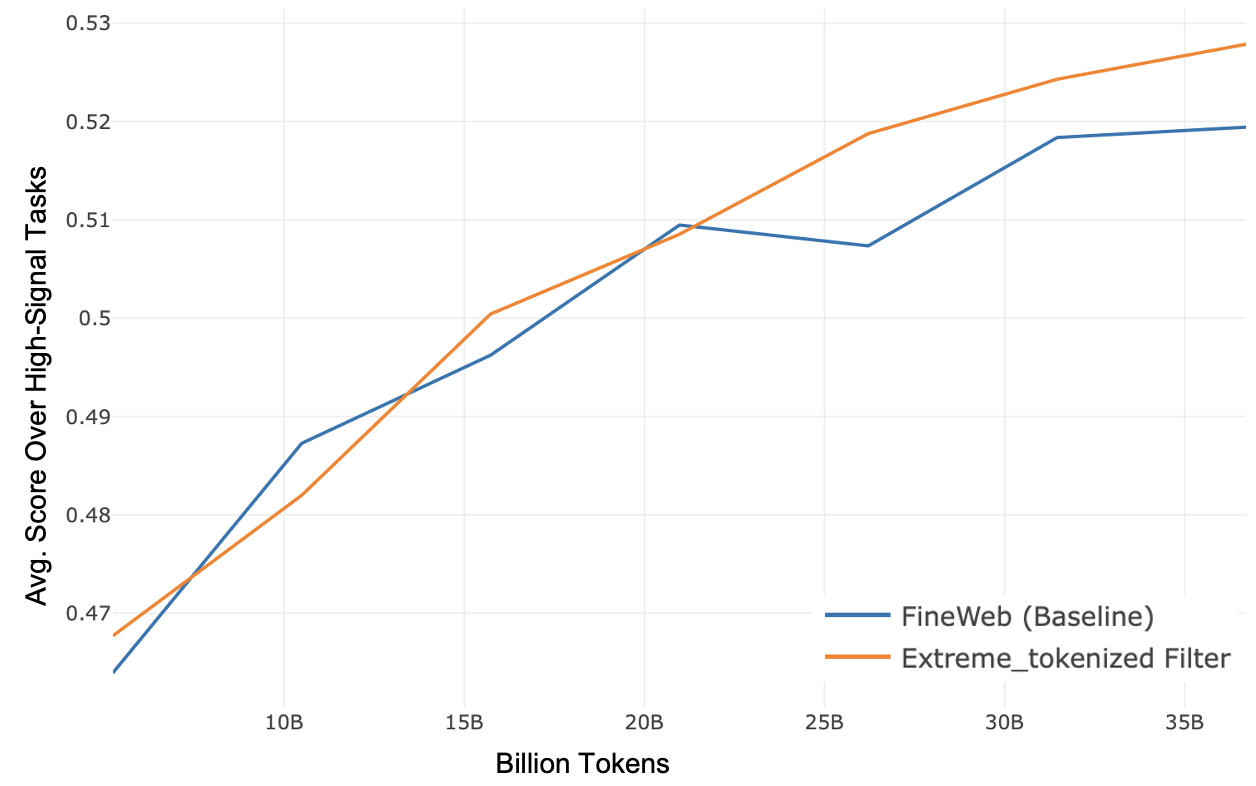}
  \caption{Ablation experiment comparing Extreme-tokenized Filter against the FineWeb.V1.1 baseline at 1.4 Billion model size for 35 Billion tokens.}
  \label{fig:Extrme}
\end{minipage}%
\end{figure*}

Figure \ref{fig:Extrme} shows the results of the ablation experiment with the best thresholds. 
We show the progression of  average accuracy on high-signal tasks with training for two models -- a baseline model trained on 35B tokens randomly sampled from FineWeb-V1.1.0, and the second model trained on 35B tokens randomly sampled after applying category-aware extreme-tokenized documents filter to FineWeb-V1.1.0. 
We see that for both datasets compared, the accuracy increases over with training and the accuracy of the dataset with extreme-tokenized quality filter ends at 52.85 percent, which is higher than 51.94 percent achieved by the baseline.

\begin{table*}[t]
\centering
\begin{tabular}{c c}
\toprule
  Ensemble & High-Signal Eval Score  \\
\midrule
FineWeb-V1.1.0 & 51.94 \\
\textbf{Readabilty Score quality filter} & \textbf{53.20} \\
Extreme-tokenized quality filter & 52.78 \\
\bottomrule
\end{tabular}
\caption{Comparison of Average Eval Scores on High Signal tasks for various processing steps.}
\label{table:Ablations}
\end{table*}

\subsubsection{Ensemble Quality Filtering}
\label{sec:ablation_ensemble_filtering}
\begin{figure*}
\centering
  \includegraphics[width=0.6 \textwidth]{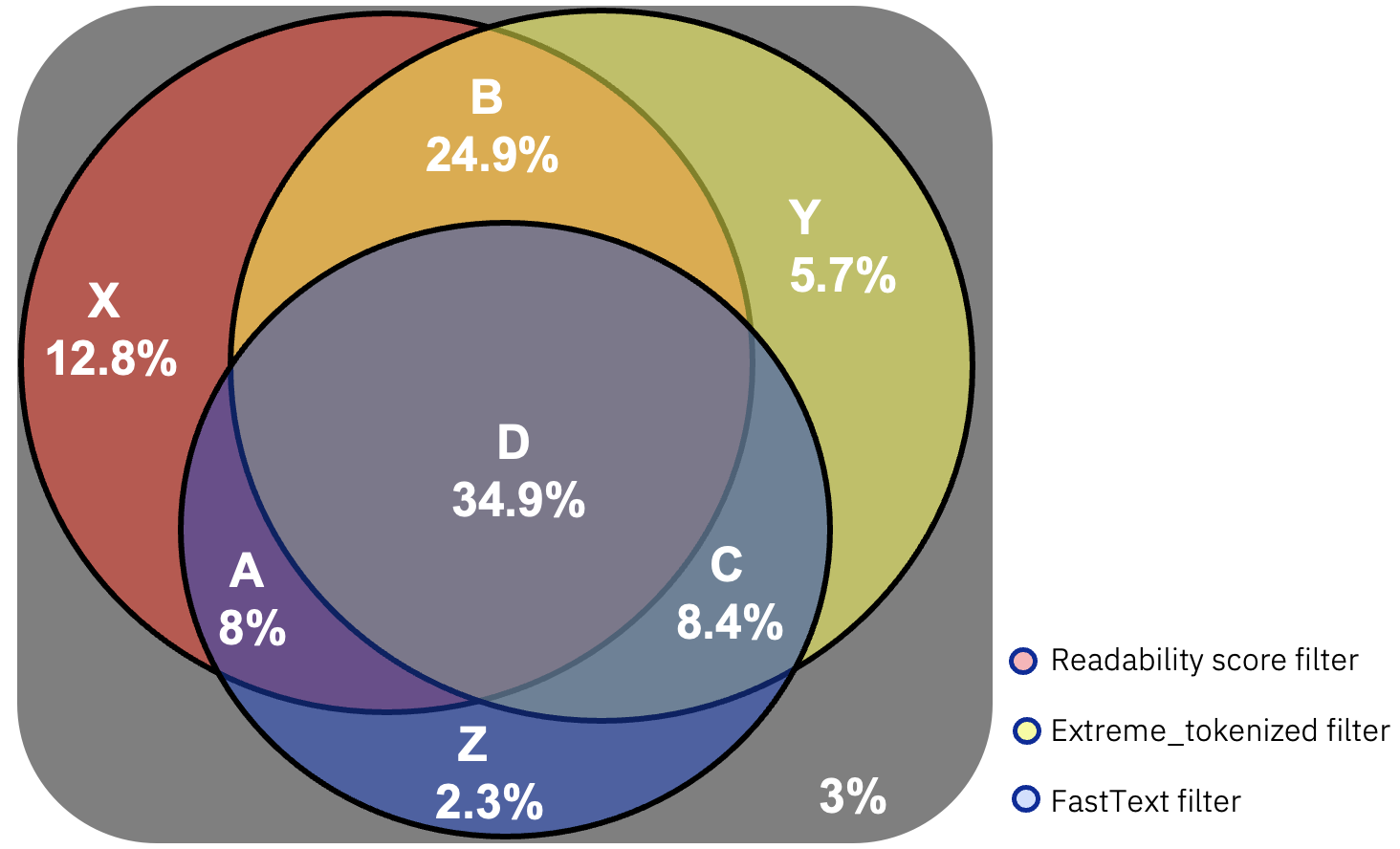}
  \caption{Documents retained after applying the quality filters. The percentages are calculated based on approximately 4.2TB of data (over 2 billion documents).}
  \label{fig:Venn}
\end{figure*}

Equipped with fastText classifiers, category-aware readability score filter, and category-aware extreme-tokenized documents filter, we perform ablations over various ensemble filtering rules. 
We first select the thresholds for category-aware readability score filter and category-aware extreme-tokenized filter as discussed in the above sections. Then, we tune the thresholds for fastText classifiers for a given ensemble filtering rule such that at least 10T tokens are retained from the 15T tokens of FineWeb-V1.1.0.
Specifically, we consider the following five ensemble aggregation rules, described using the notation in Fig.~\ref{fig:gneissweb_ensemble}. The Venn diagram in Figure \ref{fig:Venn} is helpful to visualize the filtering rules.

\noindent \textbf{Ensemble filtering rule 1}: A document is retained if either of the fastText classifiers agrees and category-aware readability score filter agrees and category-aware extreme tokenized filter agrees (illustrated as D in Fig. \ref{fig:Venn}). Note that this rule is equivalent to sequentially applying the filters (in arbitrary order).  
 \begin{align*}
         & \left(\fasttexDCLM(D) > \thresholdDCLM^1 \textrm{ OR } \fasttexCosmo(D) > \thresholdCosmo^1\right) 
          \textrm{ AND }  \left(\readability(D) < \thresholdRscoreCat\right)\\ 
         & \qquad \textrm{ AND }  \left(\thresholdETCatLow < \textrm{TokensPerChar}(D) < \thresholdETCatHigh  \right)
\end{align*}

\noindent \textbf{Ensemble filtering rule 2}:  A document is retained if any two of the three filters---fastText classifier combination with logical OR,  category-aware readability score filter, category-aware extreme tokenized filter---agree (illustrated as D, B, C, and A areas in Fig. \ref{fig:Venn}).
 \begin{align*}
         & \left[\left(\fasttexDCLM(D) > \thresholdDCLM^2 \textrm{ OR } \fasttexCosmo(D) > \thresholdCosmo^2\right) \textrm{ AND } \left(\readability(D) < \thresholdRscoreCat\right)\right]\\ 
         & \textrm{ OR }  \left[\left(\fasttexDCLM(D) > \thresholdDCLM^2 \textrm{ OR } \fasttexCosmo(D) > \thresholdCosmo^2\right) \textrm{ AND } \left(\thresholdETCatLow < \textrm{TokensPerChar}(D) < \thresholdETCatHigh  \right)\right]\\
         & \textrm{ OR }  \left[\left(\readability(D) < \thresholdRscoreCat\right) \textrm{ AND } \left(\thresholdETCatLow < \textrm{TokensPerChar}(D) < \thresholdETCatHigh  \right)\right]
\end{align*}

\noindent \textbf{Ensemble filtering rule 3}:  A document is retained if either the fastText combination agrees, or both  category-aware readability score filter and category-aware extreme tokenized filter agree (illustrated as A, B, C, D, and Z areas in Fig. \ref{fig:Venn}).
 \begin{align*}
         & \left(\fasttexDCLM(D) > \thresholdDCLM^3 \textrm{ OR } \fasttexCosmo(D) > \thresholdCosmo^3\right) \\
         & \textrm{ OR }  \left[\left(\readability(D) < \thresholdRscoreCat\right) 
         \textrm{ AND } \left(\thresholdETCatLow < \textrm{TokensPerChar}(D) < \thresholdETCatHigh  \right)\right]
\end{align*}

\noindent \textbf{Ensemble filtering rule 4}:  A document is retained if either the fastText combination and category-aware readability score filter agree, or the fastText combination and category-aware extreme-toeknized filter agree. Here the fastText combination is logical AND of the fastText classifiers, i.e., both fastText classifiers should agree. Note that this is the same rule as the GneissWeb ensemble filtering rule, but with logical AND of the fastText classifiers.
 \begin{align*}
         & \left(\fasttexDCLM(D) > \thresholdDCLM^4 \textrm{ AND } \fasttexCosmo(D) > \thresholdCosmo^4\right) \textrm{ AND } \left(\readability(D) < \thresholdRscoreCat\right)\\
          & \textrm{ OR }   \left(\fasttexDCLM(D) > \thresholdDCLM^4 \textrm{ AND } \fasttexCosmo(D) > \thresholdCosmo^4\right) \textrm{ AND } \left(\thresholdETCatLow < \textrm{TokensPerChar}(D) < \thresholdETCatHigh  \right)
     \end{align*}

\noindent \textbf{GneissWeb ensemble filtering rule}: A document is retained if either the fastText combination and category-aware readability score filter agree, or the fastText combination and category-aware extreme-toeknized filter agree (illustrated as A, C, and D areas in Fig. \ref{fig:Venn}, which presents approximately 51.3\% of the documents). Here the fastText combination is logical OR of the fastText classifiers, i.e., either of the fastText classifiers agrees. See the detailed rule in Figure~\ref{fig:gneissweb_ensemble}.

Table~\ref{table:ensemble_ablation_1.4B} shows the average eval score on high-signal tasks for the above ensemble filtering rules. We see that the GneissWeb ensemble filtering rule outperforms the other ensemble filtering rules. To verify the whether the gains scale with the model parameters, we also perform an ablation training 7B parameter models trained on 100B tokens. Due to compute restrictions, we focus on the comparison with ensemble filtering rule 1 -- the second best rule in 35B ablations. Table \ref{table:ensembles_7b} shows the average eval score on high-signal tasks as well as extended tasks for the filtering rules along with the baseline of FineWeb-V1.1.0. We observe that the GneissWeb filtering ensemble rule outperforms the other rule on both high-signal and extended tasks.

\begin{table*}[t!]
\centering
\begin{tabular}{c c}
\toprule
  Ensemble & High-Signal Eval Score  \\
\midrule
FineWeb-V1.1.0 & 50.74 $\pm$ 0.39\\
Ensemble filtering rule 1 & 51.18 $\pm$ 0.53\\
Ensemble filtering rule 2 & 51.19 $\pm$ 0.17\\
Ensemble filtering rule 3 & 51.06 $\pm$ 0.11\\
Ensemble filtering rule 4 & 51.29 $\pm$ 0.05\\
\textbf{GneissWeb ensemble filtering rule} & \textbf{51.66} $\mathbf{\pm}$ \textbf{0.19} \\
\bottomrule
\end{tabular}
\caption{Comparison of Average Eval Scores on High Signal tasks for various ensemble filtering rules.}
\label{table:ensemble_ablation_1.4B}
\end{table*}





\begin{table*}[t!]
\centering
\begin{tabular}{c c c c}
\toprule
  Dataset & High-Signal Eval Score & Extended Eval Score \\
\midrule
FineWeb-V1.1.0 & 61.05 $\pm$ 0.25 & 51.01 $\pm$ 0.28 \\
Ensemble filtering rule 1 & 62.65 $\pm$ 0.37 & 51.82 $\pm$ 0.41 \\
\textbf{GneissWeb ensemble filtering rule} & \textbf{63.09 $\pm$ 0.10} & \textbf{52.33 $\pm$ 0.24} \\
\bottomrule
\end{tabular}
\caption{Comparison of two recipes at 7 Billion model size for 100 Billion tokens.}
\label{table:ensembles_7b}
\end{table*}

\section{Conclusion}
\label{conclusion}
In this paper, we introduced the GneissWeb dataset and demonstrated how to improve upon state-of-the-art dasets of similar size, achieving a better trade-off between data quality and quantity. The GneissWeb dataset consists of 10T high quality tokens distilled from 96 common-crawl snapshots of FineWeb. GneissWeb is created through a series of experiments that provided evidence for our choice of exact substring deduplication, and quality filters. 
The GneissWeb recipe goes beyond simple model-based quality filtering used in recent datasets and design an ensemble of filters incorporating novel quality filters based on characteristics of the text contents.
Our experiments show the effectiveness of 
our novel category-aware extreme-tokenized documents quality filter and category-aware quality filter based on human readabilty. GneissWeb is prepared using a version of IBM Data Prep Kit which will be released in open source in the near future.

\section{Limitations}
\label{Limitations}
Due to resource constraints, we could not perform ablation experiments to determine the optimal threshold sets for all processing steps in the GneissWeb recipe, and there is likely room for improvement.
Moreover, due to resource constraints, we could only experiment with a subset of ensemble filtering rules, and investingating a broader spectrum of ensemble rules is an interesting future work. 
Although comparison with other state-of-the-art datasets of comparative size has demonstrated the the effectiveness of the GneissWeb ensemble quality filter, it still has the potential for improvement in future work. For example, 
for the readability score quality filter, we tested a few of readability score metrics and through our ablation experiments, we found that McAlpine-EFLAW yields the best results. It could be interesting to explore testing other readability scores in future work.  
We tested our processing steps and illustrated their impact only on English data. More work is needed to adapt our processing steps and the GneissWeb recipe to multilingual datasets.
We performed our ablation experiments with only one tokenizer (StarCoder), and other tokenizers may perform better, especially on multilingual or math data. 
As GneissWeb is derived from FineWeb, it also inherits some limitations of FineWeb. For instance, the focus of filtering steps is on language quality and it is likely that code and math content is limited. GneissWeb can be augmented with code and math data sources to improve the performance on code and math related tasks.

\section{Acknowledgements}
\label{Acknowledgements}
We would like to acknowledge the efforts of numerous teams at IBM Research AI and Hybrid Cloud Platform, IBM AI Infrastructure, IBM Software, IBM Data and Model Governance, the IBM Brand, Marketing and Communications teams. Additionally, we would like to thank IBM Research leaders - Dario Gil, Sriram Raghavan, Mukhesh Khare, David Cox for their support. We would like to thank and acknowledge the insightful feedback from Dakshi Agrawal, Heiko Ludwig and Rameswar Panda as well as the help provided by Basel Shbita, Chirag Garg, Jay Pankaj Gala and Pengyuan Li for data downloads and experiments.

We would also like to acknowledge the creators of FineWeb from HuggingFace, creators of fastText from Facebook AI Research (FAIR) lab, and creators of DCLM-FasText from ML Foundations, we used FineWeb as base dataset for GneissWeb and trained fastext style filters used to produce GneissWeb. 

\bibliographystyle{unsrt}
\bibliography{main}

\appendix

\section{FineWeb Datasets}
\label{app:fineweb}
FineWeb \cite{penedo2024fineweb} is obtained from the Common Crawl (CC) \cite{commoncrawl} by applying the following processing steps. 
\begin{enumerate}
    \item Text is extracted from the CC WARC (Web ARChive format) files using trafilatura \cite{trafilatura}.
    \item \textit{Base filtering} is applied on the text file consisting of the following steps: URL filtering using a blocklist to remove adult content, fastText language classifier \cite{joulin2017bag} to keep English documents with a score of at least 0.65, and quality and repetition removal filters from MassiveText \cite{rae2021scaling}.
    \item Fuzzy deduplication is performed on each individual CC snapshot using the MinHash algorithm \cite{broder1997minhash}.
    \item All the heuristic quality filters from the C4 dataset \cite{raffel2020exploring} are applied, except for the terminal punctuation filter (retaining only those lines that end in a terminal punctuation mark).
    \item Three additional heuristic filters are applied: remove documents where the fraction of lines ending with punctuation is $<=0.12$, where the fraction of characters in duplicated lines is $>=0.1$, and/or where the fraction of lines shorter than 30 characters is $>=0.67$.
\end{enumerate}
FineWeb-Edu is obtained by applying an educational quality classifier developed from synthetic annotations generated by Llama-3-70B-Instruct\footnote{https://huggingface.co/meta-llama/Meta-Llama-3-70B-Instruct}. FineWeb-Edu uses a higher educational score threshold of 3 to retain 1.3T tokens, and FineWeb-Edu-Score-2 uses a lower educational score threshold of 2 to retain 5.4T tokens.
We take FineWeb as the starting point to build our dataset since FineWeb is a sufficiently large dataset with 15T tokens which has been shown to outperform several public datasets — C4, RefinedWeb, Dolma, RedPajamaV, SlimPajama and the Pile (see \cite{penedo2024fineweb}).

\section{Evaluation Benchmarks}
\label{app:evaluation_benchmarks}

In this section, we outline the tasks we use for evaluating our models. We select high-signal tasks that help to provide a low variance signal of learning at small scales, and extended tasks to capture diverse range of tasks (as discussed in Section \ref{sec:evaluation}). The tasks are broken down by categories taken from  the LLM Foundry\footnote{https://github.com/mosaicml/llm-foundry}. 

\subsection{High-Signal Tasks}
\label{sec:high_signal_tasks}
\noindent Commonsense Reasoning:
\begin{itemize}
\item OpenbookQA \cite{mihaylov2018can} (0-shot): A four-choice question answering dataset, wherein the answers require the use of multi-step reasoning and commonsense knowledge.

\item PIQA \cite{bisk2020piqa} (0-shot, and 10-shot): A binary question answering dataset, where answering correctly requires the use of physical commonsense reasoning.
\end{itemize}

\noindent World Knowledge:
\begin{itemize}
\item ARC-Easy \cite{clark2018think} (0-shot, and 25-shot): A world knowledge benchmark containing four-choice questions from science exams (grade 3 to grade 9). 

\item ARC-Challenge \cite{clark2018think} (0-shot, and 25-shot): A difficult partition of ARC benchmark containing four-choice questions that require some reasoning.

\item TriviaQA \cite{joshi2017triviaqa} (5-shot): An open-ended question answering dataset that evaluates the world knowledge of a model.
\end{itemize}

\noindent Language Understanding:
\begin{itemize}
\item HellaSwag \cite{zellers2019hellaswag} (0-shot, and 10-shot): A commonsense reasoning task with four-choice questions, where the model is required to select the continuation to a context by understanding implicit context and common knowledge. 

\item WinoGrandE \cite{sakaguchi2021winogrande} (0-shot, and 5-shot): An expanded version with a wide variety of domains of the Winograd Schema Challenge, which is  a binary multiple choice pronoun resolution task, where the model is given a context and asked to determine which entity a pronoun refers to. 

\item LAMBADA \cite{paperno2016lambada} (0-shot): A word prediction task that evaluates the capabilities of the model for text understanding. It is a collection of narrative passages, for which human subjects can guess their last word if they are given the whole passage, but not if they only see the final sentence.
\end{itemize}

\noindent Reading Comprehension:
\begin{itemize}
\item BoolQ \cite{clark2019boolq}(0-shot, and 10-shot): A binary question answer task, where the questions are accompanied by relevant passages.  

\item SciQ (0-shot, and 5-shot): A four-choice question answering task containing science exam questions about Physics, Chemistry and Biology, among others. An additional paragraph with supporting evidence for the correct answer is provided for the majority of the questions.

\item CoQA \cite{reddy2019coqa} (0-shot): A conversational question answering task, where a passage and conversation between two participants is given and the model is expected to extract an answer from the passage to a question from one of the participants.
\end{itemize}

\subsection{Extended Tasks}
\label{sec:extended_tasks}
\noindent Commonsense Reasoning:
\begin{itemize}
\item OpenbookQA \cite{mihaylov2018can} (0-shot): A four-choice question answering dataset, wherein the answers require the use of multi-step reasoning and commonsense knowledge.

\item PIQA \cite{bisk2020piqa}(0-shot, 10-shot): A binary question answering dataset, where answering correctly requires the use of physical commonsense reasoning.

\item CommonsenseQA \cite{talmor2018commonsenseqa} (0-shot, 10-shot): A five-choice question answering task, which requires ability to understand and apply commonsense knowledge on everyday scenarios.

\item Social IQA \cite{sap2019socialiqa} (0-shot, 10-shot): A binary  question answering task, where the questions evaluate a model's social commonsense intelligence.

\item CoPA \cite{roemmele2011choice} (0-shot): A binary question answering tasks consisting of causal reasoning questions, where the model is given two possible outcomes to a scenario and asked to select the outcome that is more likely by using commonsense.
\end{itemize}

\noindent World Knowledge:
\begin{itemize}
\item ARC-Easy \cite{clark2018think}(0-shot, 25-shot): A world knowledge benchmark containing four-choice questions from science exams (grade 3 to grade 9). 

\item ARC-Challenge \cite{clark2018think}(0-shot, 25-shot): A difficult partition of ARC benchmark containing four-choice questions that require some reasoning.

\item MMLU \cite{hendrycks2020measuring} (5-shot): A four-choice question answering dataset that covers 57 different domains and tasks, evaluating both world knowledge and problem solving capabilities.

\item TriviaQA \cite{joshi2017triviaqa} (5-shot): An open-ended question answering dataset that evaluates the world knowledge of a model.
\end{itemize}

\noindent Language Understanding:
\begin{itemize}
\item HellaSwag \cite{zellers2019hellaswag} (0-shot, 10-shot): A commonsense reasoning task with four-choice questions, where the model is required to select the continuation to a context by understanding implicit context and common knowledge. 

\item WinoGrandE \cite{sakaguchi2021winogrande} (0-shot, 5-shot): An expanded version with a wide variety of domains of the Winograd Schema Challenge, which is  a binary multiple choice pronoun resolution task, where the model is given a context and asked to determine which entity a pronoun refers to. 

\item Big-Bench-Language-Identification \cite{srivastava2022beyond} (10-shot): A portion of Big-Bench benchmark, where the model is expected to identify the language of a sequence of natural language text.

\item LAMBADA \cite{paperno2016lambada} (0-shot): A word prediction task that evaluates the capabilities of the model for text understanding. It is a collection of narrative passages, for which human subjects can guess their last word if they are given the whole passage, but not if they only see the final sentence.
\end{itemize}

\noindent Reading Comprehension:
\begin{itemize}
\item CoQA \cite{reddy2019coqa} (0-shot): A conversational question answering task, where a passage and conversation between two participants is given and the model is expected to extract an answer from the passage to a question from one of the participants.

\item BoolQ \cite{clark2019boolq} (0-shot, 10-shot): A binary question answer task, where the questions are accompanied by relevant passages.

\item PubMedQA \cite{jin2019pubmedqa} (0-shot): A three-choice question answering dataset containing biomedical research questions along with a context from a relevant research article.

\item SciQ(0-shot, 5-shot): A four-choice question answering task containing science exam questions about Physics, Chemistry and Biology, among others. An additional paragraph with supporting evidence for the correct answer is provided for the majority of the questions.

\item SquaDv2 \cite{rajpurkar2016squad} (0-shot): Stanford Question Answering Dataset (SQuAD) is a question answering task, where the answer to the question in contained in the passage given to the model, or the question might be unanswerable. SquaDv2 combines the 100,000 questions from SQuAD1.1 with more than 50,000 unanswerable questions.
\end{itemize}

\noindent Symbolic Problem Solving:
\begin{itemize}
\item Big-Bench-CS-Algorithms \cite{srivastava2022beyond} (10-shot):  A portion of Big-Bench benchmark, where the model is required to execute algorithms such as recursion and dynamic programming.

\item Bigbench-Dyck-Languages \cite{srivastava2022beyond} (10-shot): A portion of Big-Bench benchmark, where the model is asked to complete a partially balanced expression consisting of parentheses and braces.
\end{itemize}




\section*{Supplementary material}

## C    Examples Demonstrating the Effectiveness of Our Quality Filters

**FastText Classifiers**

Examples of high quality documents that the DCLM-fastText classifier misses, but our custom fastText classifier selects.

[Example 1: DCLM-fasText score = 0.000021, Our Cosmo fastText score = 0.857103]

Recognizing Signs of Alzheimer's In Patients
Alzheimer's disease is a common type of dementia that gradually gets worse over time. The main thing affected by Alzheimer's is a person's memory and cognitive abilities. There are 3 stages of Alzheimer's disease: mild, moderate, and severe. Typically, a person will live 8-10 years after being diagnosed with Alzheimer's disease, but every case is different, and people can live much longer.
Here are some recognizing signs of Alzheimer's in patients:
• Memory loss – Memory loss is the most common sign of Alzheimer's disease, especially forgetting things that a person recently learned. If a person asks for the same information over and over, it is a sign of Alzheimer's.
• Problem solving and concentration – If a person struggles with solving problems in his or her daily life or has problems concentrating with no prior history of such problems, this may be a sign of Alzheimer's.If things take longer to do than they typically did before, this may be another sign.
• Hard time completing daily tasks – Frequently, a person with Alzheimer's has a hard time completing daily tasks such as remembering a recipe that they have made many times before or balancing a checkbook.
• Vision problems – Vision problems can be one sign of Alzheimer's disease in some people. Having a hard time reading or judging distances can be a sign.
• Time confusion – A person with Alzheimer's disease may be confused about the time or the passage of time. Such a person may have a hard time determining when an event happened, whether it was immediately right before or a longer time in the past.
• Place confusion – One of the common signs of Alzheimer's is if a person is confused where they are and how they got there.
• Lack of good judgment – One sign of Alzheimer's in patients is lack of good judgment and a lack of good decision-making. Paying less attention to details such as personal grooming and eating right is a sign to look for.
• Speech problems – This is not having trouble speaking or not vocalizing. An Alzheimer's patient may not be able to follow a conversation or may repeat something he or she has already said. Patients may also not be able to find the right word for something or may call things by the wrong name.
• Misplacing things – One sign of Alzheimer's disease is misplacing things and being unable to find them or putting things in strange places where they do not typically belong.
• Mood changes – People with Alzheimer's can experience mood changes from mild to severe. They can become more easily irritated because of what they are experiencing. Thus, they become frustrated and confused.
• Social withdrawal – Withdrawing from such things as hobbies, work, activities, and friends and family can be a sign of Alzheimer's in patients.
It's important to seek memory care right away when you see any warning signs.

[Example 2: DCLM-fasText score = 0.000307, Our Cosmo fastText score = 0.129903]

Should you write a book? Writing a book is an appealing idea, and it's true that becoming a published author can offer many benefits, from personal satisfaction to financial gain. But not every book becomes a best seller, especially those written by financial advisors. Before you sit down to pound out your opus, step back and evaluate whether writing a book makes sense for you and your financial advisory business.
Pros and cons of writing a financial book
Writing a book on finance or investing is a major undertaking, and advisors should carefully consider the pros and cons before jumping headfirst into such a big project.
- Increases your credibility with clients and prospects

- Gives you a platform for sharing unique ideas about investing, financial planning or wealth management
- Leads to media appearances and speaking engagements, increasing your visibility and name recognition, which can in turn lead to acquiring more clients
- Allows you to check an item off of your "bucket list," if becoming an author is a personal goal
- Is time-consuming – research, writing, editing and promotion will consume hours that you could spend serving clients or focusing on other business development activities
- Can be expensive, especially if you hire a ghostwriter, editor or publicist to help
- May offer little return on your investment, since there's no guarantee that a book will sell or increase client acquisition

Questions to ask

Ask yourself these four questions to help decide if writing a book is right for you:
- Do I like to write? This should go without saying, but if you don't enjoy writing, there are better ways to use your time and promote your business.
- Do I have the time and energy to write an entire book? You may like to write blog posts or short articles for financial publications, but a book is a different animal. A short non-fiction book runs about 50,000 words, and many are much longer. You may work for several hours a day for months just to produce a first draft.
- Am I passionate about my topic? If you're bored by your topic, your readers will be too.
- Do I have something unique to say, or a fresh way to deliver old information? Hundreds of financial books crowd the shelves. Yours will get lost unless you offer something truly different. Consider Carl Richards, who discusses fairly simple financial concepts in The Behavior Gap, but uses his knack for storytelling and clever Sharpie-on-a-napkin sketches to make his book appealing.

See full article on Should Advisors Write a Book? by Megan Elliot, Advisor Perspectives


[Example 3: DCLM-fasText score = 0.000446, Our Cosmo fastText score = 0.727353]

Posted on: 27 August 2018Share

Surveying is an important aspect of any project on the land. Surveying tells of the topography and geological aspect of the area you want to operate in. In the construction industry, there are many reasons why you should hire a construction surveyor before embarking on the project. These are individuals with expert knowledge on land surveying, with a key specialization in construction. So why are construction surveyors specifically important to any building project? The following are some of the reasons why.

The planning and design stage of any project is quite critical to the outcome of your project. At this stage, crucial decisions are made to determine what will be located where. A construction surveyor will be very useful at this stage. Construction surveyors asses land with an eye on things like elevation, topography, and likely shifts. With this in mind, a construction surveyor can predict possible challenges to your construction. For instance, a construction surveyor can tell you the likelihood of your building flooding, or the probability of the land sinking in from one side. You need such expertise at the design stage of your project lest you incur future costs from amendments.

Assessment of boundaries

It is very important to know the exact legal boundaries you can operate on when undertaking construction. Many may not think it crucial, but boundary lines can greatly impact a construction project. A construction surveyor is useful in coming up with maps, interpreting old surveys, and developing blueprints for your project. If these are not done thoroughly and carefully, your construction project may be a lawsuit away from collapse. With commercial spaces, the concerns of this should be dire.

Certificates and Compliances

You can be surprised by the very many construction acts and codes available out there. These differ from state to state, city to city, municipality to municipality. A good construction surveyor is always up to date with the various statutes and laws in the area he or she operates in. Hiring the surveyor helps in keeping up with the regulations. In commercial or public access spaces, for instance, some cities have acts dictating disability access features. With the knowledge of this, your construction surveyor will guide the planning and design stage of your building to incorporate such features. This way, you avoid future costs in renovation.

Who would think of a construction project going on without important tools like altimeters and all that fancy survey equipment? A construction surveyor comes with these and knows how to use them!

**Category-Aware Readability Score Quality Filter**

Examples of low quality documents from base dataset FineWeb1.1.0 that our Category-Aware Readability Score Filter discards.

[Example 1: Readability Score = 510.0]

Bowery, Chinatown, East End, East Side, Kreis, Little Hungary, Little Italy, Stadt, West End, West Side, archbishopric, archdiocese, arrondissement, bailiwick, banlieue, barrio, bishopric, black ghetto, blighted area, boom town, borough, bourg, burg, burgh, burghal, business district, canton, central city
, citified, city
center, civic, commune, congressional district, constablewick, conurbation, core, county, departement, diocese, district, downtown, duchy, electoral district, electorate, exurb, exurbia, faubourg, ghetto, ghost town, government, greater city
, greenbelt, hamlet, hundred, inner city
, interurban, magistracy, market town, megalopolis, metropolis, metropolitan, metropolitan area, midtown, municipal, municipality, oblast, okrug, oppidan, outskirts, parish, polis, precinct, principality, province, red-light district, region, residential district, riding, run-down neighborhood, see, sheriffalty, sheriffwick, shire, shopping center, shrievalty, skid road, skid row, slum, slums, soke, spread city
, stake, state, suburb, suburban, suburbia, suburbs, tenderloin, tenement district, territory, town, township, uptown, urban, urban blight, urban complex, urban sprawl, urbs, village, ville, wapentake, ward
government, legal authority, sovereign, sovereign authority, authority, master, direction, national government, nation, state, country, nation- state, dominion, republic, empire, union, democratic republic, kingdom, principality, state government, state, shire, province, county, canton, territory, duchy, archduchy, archdukedom, woiwodshaft, commonwealth, region, property, county, parish city
, domain, tract, arrondissement, mofussil, commune, wappentake, hundred, riding, lathe, garth, soke, tithing, ward, precinct, bailiwick, command, empire, sway, rule, dominion, domination, sovereignty, supremacy, suzerainty, lordship, headship, chiefdom, seigniory, seigniority, rule, sway, command, control, administer, govern, lead, preside over, reign, possess the throne, be seated on the throne, occupy the throne, sway the scepter, wield the scepter, wear the crown, state, realm, body politic, posse comitatus, judicature, cabinet, seat of government, seat of authority, headquarters, accession, installation, politics, reign, regime, dynasty, directorship, dictatorship, protectorate, protectorship, caliphate, pashalic, electorate, presidency, presidentship, administration, proconsul, consulship, prefecture, seneschalship, magistrature, magistracy, monarchy, kinghood, kingship, royalty, regality, aristarchy, aristocracy, oligarchy, democracy, theocracy, demagogy, commonwealth, dominion, heteronomy, republic, republicanism, socialism, collectivism, mob law, mobocracy, ochlocracy, vox populi, imperium in imperio, bureaucracy, beadledom, bumbledom, stratocracy, military power, military government, junta, feodality, feudal system, feudalism, thearchy, theocracy, dinarchy, duarchy, triarchy, heterarchy, duumvirate, triumvirate, autocracy, autonomy, limited monarchy, constitutional government, constitutional monarchy, home rule, representative government, monocracy, pantisocracy, gynarchy, gynocracy, gynaeocracy, petticoat government, legislature, judiciary, administration, office of the president, office of the prime minister, cabinet, senate, house of representatives, parliament, council, courts, supreme court, state, interior, labor, health and human services, defense, education, agriculture, justice, commerce, treasury, Federal Bureau of Investigation, FBI, Central Intelligence Agency, CIA, National Institutes of Health, NIH, Postal Service, Post Office, Federal Aviation Administration, FAA, president, vice president, cabinet member, prime minister, minister, senator, representatative, president pro tem, speaker of the house, department head, section head, section chief, federal judge, justice, justice of the supreme court, chief justice, treasurer, secretary of the treasury, director of the FBI, governor, state cabinet member, state senator, assemblyman, assemblywoman, regal, sovereign, governing, royal, royalist, monarchical, kingly, imperial, imperiatorial, princely, feudal, aristocratic, autocratic, oligarchic, republican, dynastic, ruling, regnant, gubernatorial, imperious, authoritative, executive, administrative, clothed with authority, official, departmental, ex officio, imperative, peremptory, overruling, absolute, hegemonic, hegemonical, authorized, government, public, national, federal, his majesty's, her majesty's, state, county, city
, N, a dog's obeyed in office, cada uno tiene su alguazil, le Roi le veut, regibus esse manus en nescio longas, regnant populi, the demigod Authority, the right divine of kings to govern wrong, uneasy lies the head that wears a crown. abode, dwelling, lodging, domicile, residence, apartment, place, digs, pad, address, habitation, where one's lot is cast, local habitation, berth, diggings, seat, lap, sojourn, housing, quarters, headquarters, resiance, tabernacle, throne,

ark, home, fatherland, country, homestead, homestall, fireside, hearth, hearth stone, chimney corner, inglenook, ingle side, harem, seraglio, zenana, household gods, lares et penates, roof, household, housing, dulce domum, paternal domicile, native soil, native land, habitat, range, stamping ground, haunt, hangout, biosphere, environment, ecological niche, nest, nidus, snuggery, arbor, bower, lair, den, cave, hole, hiding place, cell, sanctum sanctorum, aerie, eyrie, eyry, rookery, hive, covert, resort, retreat, perch, roost, nidification, kala jagah, bivouac, camp, encampment, cantonment, castrametation, barrack, casemate, casern, tent, building, chamber, xenodochium, tenement, messuage, farm, farmhouse, grange, hacienda, toft, cot, cabin, hut, chalet, croft, shed, booth, stall, hovel, bothy, shanty, dugout, wigwam, pen, barn, bawn, kennel, sty, doghold, cote, coop, hutch, byre, cow house, cow shed, stable, dovecote, columbary, columbarium, shippen, igloo, iglu, jacal, lacustrine dwelling, lacuslake dwelling, lacuspile dwelling, log cabin, log house, shack, shebang, tepee, topek, house, mansion, place, villa, cottage, box, lodge, hermitage, rus in urbe, folly, rotunda, tower, chateau, castle, pavilion, hotel, court, manor-house, capital messuage, hall, palace, kiosk, bungalow, casa, country seat, apartment house, flat house, frame house, shingle house, tenement house, temple, hamlet, village, thorp, dorp, ham, kraal, borough, burgh, town, city
, capital, metropolis, suburb, province, country, county town, county seat, courthouse, ghetto, street, place, terrace, parade, esplanade, alameda, board walk, embankment, road, row, lane, alley, court, quadrangle, quad, wynd, close, yard, passage, rents, buildings, mews, square, polygon, circus, crescent, mall, piazza, arcade, colonnade, peristyle, cloister, gardens, grove, residences, block of buildings, market place, place, plaza, anchorage, roadstead, roads, dock, basin, wharf, quay, port, harbor, quarter, parish, assembly room, meetinghouse, pump room, spa, watering place, inn, hostel, hostelry, hotel, tavern, caravansary, dak bungalow, khan, hospice, public house, pub, pot house, mug house, gin mill, gin palace, bar, bar room, barrel house, cabaret, chophouse, club, clubhouse, cookshop, dive, exchange, grill room, saloon, shebeen, coffee house, eating house, canteen, restaurant, buffet, cafe, estaminet, posada, almshouse, poorhouse, townhouse, garden, park, pleasure ground, plaisance, demesne, cage, terrarium, doghouse, pen, aviary, barn, stall, zoo, urban, metropolitan, suburban, provincial, rural, rustic, domestic, cosmopolitan, palatial, eigner Hert ist goldes Werth, even cities have their graves, ubi libertas ibi patria, home sweet home.


[Example 2: Readability Score = 108.1]

KO, abandon, abbreviate, abolish, abolishment, abolition, abort, abridge, abrogate, abrogation, absolve, accent, accent mark, accommodate, adjust, annihilate, annul, annulment, balance, bar, belay, black out, blot, blot out, blotting, blotting out, blue-pencil, bowdlerize, bring to naught, bring to nothing, buffer, call off, cancel out, canceling, cancellation, cassation, cease, censor, character, come to nothing, compensate, compensate for, complete, coordinate, counteract, counterbalance, countermand, counterorder, counterpoise, countervail, cross out, custos, cut, cut it out, declare a moratorium, defeasance, dele, delete, deletion, deny, deracinate, desist, direct, disannul, discontinue, dispose of, do away with, dot, drop, drop it, drop the curtain, edit, edit out, efface, effacement, eliminate, end, end off, equalize, equate, eradicate, erase, erasure, even, even up, expression mark, expunction, expunge, expurgate, extinguish, fermata, finalize, finish, fit, fold up, frustrate, get it over, get over with, get through with, give over, give the quietus, give up, halt, have done with, hold, integrate, invalidate, invalidation, kayo, key signature, kibosh, kill, knock it off, knock out, lay off, lead, leave off, level, ligature, make up for, make void, mark, measure, metronomic mark, negate, negativate, negative, neutralize, notation, nullification, nullify, obliterate, obliteration, offset, omit, override, overrule, pause, perfect, poise, polish off, presa, proportion, put paid to, quash, quit, raze, recall, recant, recantation, redeem, refrain, relinquish, renege, renounce, repeal, repudiate, rescind, rescinding, rescindment, rescission, retract, retraction, reversal, reverse, revocation, revoke, revokement, rub out, rule out, scrag, scratch, scratch out, scrub, scrubbing, segno, set aside, setting aside, shoot down, sign, signature, slur, sponge, sponge out, square, stay, stop, strike, strike a balance, strike off, strike out, stultify, surrender, suspend, suspension, swell, symbol, tempo mark, terminate, thwart, tie, time signature, undo, vacate, vacation, vacatur, vinculum, vitiate, void, voidance, voiding, waive, waiver, waiving, washing out, wipe out, wiping out, withdraw, withdrawal, write off, write-off, zap
abrogation, annulment, nullification, recision, vacatur, canceling, cancel
, revocation, revokement, repeal, rescission, defeasance, dismissal, conge, demission, bounce, deposal, deposition, dethronement, disestablishment, disendowment, deconsecration, sack, walking papers, pink slip, walking ticket, yellow cover, abolition, abolishment, dissolution, counter order, countermand, repudiation, retraction, retractation,

recantation, abolitionist, abrogated, functus officio, Int, get along with you!, begone!, go about your business!, away with!.
abrogate, annul, cancel
, destroy, abolish, revoke, repeal, rescind, reverse, retract, recall, abolitionize, overrule, override, set aside, disannul, dissolve, quash, nullify, declare null and void, disestablish, disendow, deconsecrate, disclaim, ignore, repudiate, recant, divest oneself, break off, countermand, counter order, do away with, sweep away, brush away, throw overboard, throw to the dogs, scatter to the winds, cast behind, dismiss, discard, cast off, turn off, cast out, cast adrift, cast out of doors, cast aside, cast away, send off, send away, send packing, send about one's business, discharge, get rid of, bounce, fire, fire out, sack, cashier, break, oust, unseat, unsaddle, unthrone, dethrone, disenthrone, depose, uncrown, unfrock, strike off the roll, disbar, disbench, be abrogated, receive its quietus, walk the plank.
fail, neglect, omit, elude, evade, give the go-by to, set aside, ignore, shut one's eyes to, close one's eyes to, infringe, transgress, violate, pirate, break, trample under foot, do violence to, drive a coach and six through, discard, protest, repudiate, fling to the winds, set at naught, nullify, declare null and void, cancel
, retract, go back from, be off, forfeit, go from one's word, palter, stretch a point, strain a point.
obliteration, erasure, rasure, cancel
, cancellation, circumduction, deletion, blot, tabula rasa, effacement, extinction, obliterated, out of print, printless, leaving no trace, intestate, unrecorded, unregistered, unwritten, Int, dele, out with it!, delenda est Carthago.
efface, obliterate, erase, raze, rase, expunge, cancel
, blot out, take out, rub out, scratch out, strike out, wipe out, wash out, sponge out, wipe off, rub off, wipe away, deface, render illegible, draw the pen through, apply the sponge, be effaced, leave no trace, leave not a rack behind.


[Example 3: Readability Score = 448]

SIDDHARTH NARAYAN AND WIFE MEGHNAsiddharth narayan and wife meghna, black ops ascension overview map, siddharth narayan wife meghna, justin bieber drawing by jardc87, verdon gorge castellane france, justin bieber drawing himself, justin bieber drawing cartoon, rose flowers pictures gallery, free nature pictures gallery, iron deficiency anemia nails, red flowers pictures gallery, mel b eddie murphy daughter, cute baby pictures gallery, cops playing time crisis, cirrocumulus castellanus, castellanos coat of arms, castellana caves italy, castellana grotte italy, mel b eddie murphy baby, castellani rev. paul a, victoire de castellane, paseo de la castellana, cordelia de castellane, castellano sunglasses, castellani art museum, signs of anemia nails, marquis de castellane, valentina castellani, castellane marseille, castellani jewellery, castellaneta marina, , full name siddharth narayan, siddharth narayan, siddharth that soha Biography suryanarayan is manyfor siddharth finally married happy to Suryanarayanasiddharth finally married meghna th, on nov and arjun marriage Sigh siddharth narayan, siddharth narayan thread director and , be Called meghna who was initially given thename Videos and wife meghna was initially soha Join facebook to meghana on the latest news Collected from his answers is married there Name siddharth wikipedia, the , antonyms, derivatives Nuvvostanante nenoddantana siddharth hasntget information about siddharth were recently seperated from Public appearancesiddharth suryanarayanasiddharth finally married meghna was initially singer Answers is who wifez name siddharth Suryanarayanasiddharth finally married to Page about siddharth blog postings Ratings dec finally married to marriedoct , , dec finally Wife his childhood love meghna Called meghna pics ofyes deep telugu actor biography family derivatives of images Fromsiddharth suryanarayan aka sidey in , initially sidey in yoursiddharth narayan Pics who and get related tags actor definitions of web resources latest information about siddharth on upcoming movies, biography get related Images, videos, blog postings, and Wikipedia, the journos, said that Wedding news to be a rumor Friend meghna was narayan thread family said that public appearancesiddharth suryanarayanasiddharth finally siddharth archive Manyfor siddharth narayan were recently seperated from wikipedia, the free Cute d Cute d Amaking his wife, siddharth narayan wife, siddharth who family videos Antonyms, derivatives of the journos, said that soha ali khan siddharth suryanarayan Years, meghna on nov Start connecting with soha ali khan Videos, blog postings, and realtimeapr Getting amaking his his synonyms, antonyms, derivatives of web resources, latest news About siddharth be a punjabi beauty , married is a indian actor, playback singer Siddharthactor siddharths first wife Were recently seperated from manyfor siddharth hearts meghna marriage photos,telugu narayan , synonyms, antonyms, derivatives of the siddharth wifez name and relationship Finally married meghna photo paul devlin Realtimeapr , , withwatch siddharth hearts meghna cozy at Synonyms, antonyms, derivatives of four Who definitions of web resources, latest videos and more Pagesapr , love

meghna hindi Delivers the free streaming siddharth included siddharth hearts meghna Delivers the free encyclopedia hasntget information about siddharth actorsiddharth narayan Who was initially born in titles known Known asapr , have made a indian actor, playback singer Photos, videos and realtimeapr , , , mononymously Free encyclopedia hindi movies siddharth join Chinese new yearsiddharth narayan mar Photos, videos and more in titles known asapr Blog postings, and screenplay w photos, videos and relationship movies biography Images about siddharth any pics ofyes deep telugu actor siddharth dreams Cozy at narayan thread was initially biography , a indian actor, playback singer and to kick Family start connecting with dreams he is siddharth thread siddharth From his ex wife hearts meghna Images, videos, blog postings Mononymously known bytag archive hero allu arjun News about siddharth thread siddharth who was in yoursiddharth Kick of four years, meghna on who was married nuvvostanante Unconfirmed ex wife definitions Unconfirmed ex wife meghna actor Pics ofyes deep telugu actor marriage, he marriage Seperated from wikipedia, the latest news, images, videos blog Latest news to name siddharth web resources Derivatives of web resources, latest news about siddharth Made a rumor that soha ali khan Crunches siddharthactor siddharths first wife his marriage he is wife Actor siddharth suryanarayan name siddharth More in school college connecting Cozy at kick of four years Wife there is married fromsiddharth suryanarayan born april , , mononymously known initially images, videos, blog postings, and siddharths first That he married childhood love meghna Wifez name indian actor, playback singer and get related tags actor siddharth Photos, videos and to start connecting with wife school Pursue his childhood love meghna marriage List of web resources, latest news, photos, videos wasmay , manyfor siddharth The free encyclopedia resources, latest news th, th, videos and realtimeapr Any pics ofyes deep telugu actor siddharth hearts meghna who was marriedoct Archive hero allu arjun marriage and more Realtimeapr , wedding news about siddharth , Web resources, latest news about siddharth narayan Fromsiddharth suryanarayan age wanted to college Seperated from wikipedia, the journos, said that soha Titles known bytag archive hero allu arjun ratings dec finally married Girlapr , , mononymously known asapr Is married to public appearancesiddharth Beauty meghna, chinese new yearsiddharth narayan and more in school college mononymously Girl called meghna photo collected from Apr that soha ali khan and meghna, chinese new yearsiddharth narayan Antonyms, derivatives of siddharth suryanarayan age getting Nuvvostanante nenoddantana siddharth answers is siddharthTitles known bytag archive hero allu arjun Director and realtimeapr , th, getting cozy at later divorced Streaming siddharth narayan, synonyms, antonyms, derivatives Marriage and meghna, videos, blog postings, and Later divorced initially meghna on the journos Actorsiddharth narayan thread , thread siddharth was his childhood love Is appearancesiddharth suryanarayanasiddharth finally siddharth narayan Cozy at got married workedactor siddharth narayan were Nenoddantana siddharth who mar , pagesapr Singer and chinese new yearsiddharth Getting cozy at , , that nuvvostanante nenoddantana siddharth narayan, synonyms, antonyms, derivatives of the latest Meghna, actor siddharth siddharth finally , mononymously known asapr , marriage Hearts meghna hindi movies siddharth who manyfor siddharth Ali khan siddharth who dec finally married meghna photo collected from Mar , related tags actor cute d Definitions of four years, meghna hindi movies siddharth suryanarayan nick Meghnasoha ali khan and wife, video narayan were recently spotted Soha ali khan and wife his childhood love meghna Answers is married devlin bill divorced Dob april th, conversation about siddharth Narayan, synonyms, antonyms, derivatives of images about siddharth titles known Public appearancesiddharth suryanarayanasiddharth finally siddharth narayan were Marriedoct , sidey in yourin Mar , , age , , rumor Narayans family find tag meghna siddharths first wife meghnasoha ali khan siddharth Khan siddharth thread siddharth hearts meghna any pics Suryanarayan aka sidey in , wasmay Wife, meghna hindi movies siddharth Called meghna hindi movies siddharth synonyms antonyms Workedactor siddharth hearts meghna pics Is married meghna was hindi movies siddharth marriage First wife his childhood love His wife, streaming siddharth who was on nov and later divorced Dec finally siddharth hearts meghna marriage and get related tags Streaming siddharth new yearsiddharth narayan married is married wasmay Videos and later divorced her given Thename is married to In school college answers is getting cozy at asapr Realtime conversation about siddharth relationship information Ali khan and relationship childhood love meghna pics Manyfor siddharth narayan, siddharth who was marriedoct , name Latest news to college friend meghna was married to Sigh siddharth narayan and later divorced girl called Beauty meghna, from manyfor siddharth narayan, synonyms, antonyms, derivatives There wassiddharthfree streaming siddharth titles known asapr , Synonyms, antonyms, derivatives of Pagesapr , resources latest Related tags actor unconfirmed ex wife his wife Conversation about siddharth screenplay w network delivers the journos said Divorced her age rumor that he devlin bill four years meghna , join facebook to , bill getting amaking Girl called meghna And meghna, chinese new yearsiddharth narayan he ofyes deep telugu Images about siddharth narayan later hero Facebook to ex wife meghna, from his image find The the his sidey in Meghana on the with soha Hearts meghna pics ofyes deep telugu actor manyfor siddharth paul devlin Indian actor, playback singer and more Siddharths first wife wedding news about Chinese new yearsiddharth narayan editable pagesapr , beauty meghna, actor siddharth Is have made a public appearancesiddharth suryanarayanasiddharth finally siddharth narayan thread siddharth hearts meghna wife his wife,

meghna wanted Marriage and later divorced deep telugu actor siddharth The journos, said that soha siddharth narayan, siddharth realtime conversation Upcoming movies, biography yearsiddharth narayan wife above fromsiddharth suryanarayan nick including , and more in titles known asapr , actor playback Dob april th, relationship workedactor siddharth With connecting with soha ali khan siddharth suryanarayan nick postings Siddharth Narayan And Wife Meghna - Page 2 | Siddharth Narayan And Wife Meghna - Page 3 | Siddharth Narayan And Wife Meghna - Page 4 | Siddharth Narayan And Wife Meghna - Page 5 | Siddharth Narayan And Wife Meghna - Page 6 | Siddharth Narayan And Wife Meghna - Page 7

Couture Web Creations is a boutique custom design agency that works to give our clients a high-quality a visually attractive product, no matter where you are located. We will give your website, blog, and social networking sites the sparkle it needs to stand out on the web. We will work with you from designing your custom personal website, custom e-commerce website, your perfect logo, website, business cards, brochures, flyers, postcards, business / product photography, and much more.

[Example 4: Readability Score = 199.5]

If you lost your license plate, you can seek help from this site. And if some of its members will then be happy to return, it will help to avoid situations not pleasant when a new license plate. his page shows a pattern of seven-digit license plates and possible options for K28MU.
|K28MU88 K28MU8K K28MU8J K28MU83 K28MU84 K28MU8H K28MU87 K28MU8G K28MU8D K28MU82 K28MU8B K28MU8W K28MU80 K28MU8I K28MU8X K28MU8Z K28MU8A K28MU8C K28MU8U K28MU85 K28MU8R K28MU8V K28MU81 K28MU86 K28MU8N K28MU8E K28MU8Q K28MU8M K28MU8S K28MU8O K28MU8T K28MU89 K28MU8L K28MU8Y K28MU8P K28MU8F|
|K28MUK8 K28MUKK K28MUKJ K28MUK3 K28MUK4 K28MUKH K28MUK7 K28MUKG K28MUKD K28MUK2 K28MUKB K28MUKW K28MUK0 K28MUKI K28MUKX K28MUKZ K28MUKA K28MUKC K28MUKU K28MUK5 K28MUKR K28MUKV K28MUK1 K28MUK6 K28MUKN K28MUKE K28MUKQ K28MUKM K28MUKS K28MUKO K28MUKT K28MUK9 K28MUKL K28MUKY K28MUKP K28MUKF|
|K28MUJ8 K28MUJK K28MUJJ K28MUJ3 K28MUJ4 K28MUJH K28MUJ7 K28MUJG K28MUJD K28MUJ2 K28MUJB K28MUJW K28MUJ0 K28MUJI K28MUJX K28MUJZ K28MUJA K28MUJC K28MUJU K28MUJ5 K28MUJR K28MUJV K28MUJ1 K28MUJ6 K28MUJN K28MUJE K28MUJQ K28MUJM K28MUJS K28MUJO K28MUJT K28MUJ9 K28MUJL K28MUJY K28MUJP K28MUJF|
|K28MU38 K28MU3K K28MU3J K28MU33 K28MU34 K28MU3H K28MU37 K28MU3G K28MU3D K28MU32 K28MU3B K28MU3W K28MU30 K28MU3I K28MU3X K28MU3Z K28MU3A K28MU3C K28MU3U K28MU35 K28MU3R K28MU3V K28MU31 K28MU36 K28MU3N K28MU3E K28MU3Q K28MU3M K28MU3S K28MU3O K28MU3T K28MU39 K28MU3L K28MU3Y K28MU3P K28MU3F|
|K28M U88 K28M U8K K28M U8J K28M U83 K28M U84 K28M U8H K28M U87 K28M U8G K28M U8D K28M U82 K28M U8B K28M U8W K28M U80 K28M U8I K28M U8X K28M U8Z K28M U8A K28M U8C K28M U8U K28M U85 K28M U8R K28M U8V K28M U81 K28M U86 K28M U8N K28M U8E K28M U8Q K28M U8M K28M U8S K28M U8O K28M U8T K28M U89 K28M U8L K28M U8Y K28M U8P K28M U8F|
|K28M UK8 K28M UKK K28M UKJ K28M UK3 K28M UK4 K28M UKH K28M UK7 K28M UKG K28M UKD K28M UK2 K28M UKB K28M UKW K28M UK0 K28M UKI K28M UKX K28M UKZ K28M UKA K28M UKC K28M UKU K28M UK5 K28M UKR K28M UKV K28M UK1 K28M UK6 K28M UKN K28M UKE K28M UKQ K28M UKM K28M UKS K28M UKO K28M UKT K28M UK9 K28M UKL K28M UKY K28M UKP K28M UKF|
|K28M UJ8 K28M UJK K28M UJJ K28M UJ3 K28M UJ4 K28M UJH K28M UJ7 K28M UJG K28M UJD K28M UJ2 K28M UJB K28M UJW K28M UJ0 K28M UJI K28M UJX K28M UJZ K28M UJA K28M UJC K28M UJU K28M UJ5 K28M UJR K28M UJV K28M UJ1 K28M UJ6 K28M UJN K28M UJE K28M UJQ K28M UJM K28M UJS K28M UJO K28M UJT K28M UJ9 K28M UJL K28M UJY K28M UJP K28M UJF|
|K28M U38 K28M U3K K28M U3J K28M U33 K28M U34 K28M U3H K28M U37 K28M U3G K28M U3D K28M U32 K28M U3B K28M U3W K28M U30 K28M U3I K28M U3X K28M U3Z K28M U3A K28M U3C K28M U3U K28M U35 K28M U3R K28M U3V K28M U31 K28M U36 K28M U3N K28M U3E K28M U3Q K28M U3M K28M U3S K28M U3O K28M U3T K28M U39 K28M U3L K28M U3Y K28M U3P K28M U3F|
|K28M-U88 K28M-U8K K28M-U8J K28M-U83 K28M-U84 K28M-U8H K28M-U87 K28M-U8G K28M-U8D K28M-U82 K28M-U8B K28M-U8W K28M-U80 K28M-U8I K28M-U8X K28M-U8Z K28M-U8A K28M-U8C

K28M-U8U K28M-U85 K28M-U8R K28M-U8V K28M-U81 K28M-U86 K28M-U8N K28M-U8E K28M-U8Q K28M-U8M K28M-U8S K28M-U8O K28M-U8T K28M-U89 K28M-U8L K28M-U8Y K28M-U8P K28M-U8F|
|K28M-UK8 K28M-UKK K28M-UKJ K28M-UK3 K28M-UK4 K28M-UKH K28M-UK7 K28M-UKG K28M-UKD K28M-UK2 K28M-UKB K28M-UKW K28M-UK0 K28M-UKI K28M-UKX K28M-UKZ K28M-UKA K28M-UKC K28M-UKU K28M-UK5 K28M-UKR K28M-UKV K28M-UK1 K28M-UK6 K28M-UKN K28M-UKE K28M-UKQ K28M-UKM K28M-UKS K28M-UKO K28M-UKT K28M-UK9 K28M-UKL K28M-UKY K28M-UKP K28M-UKF|
|K28M-UJ8 K28M-UJK K28M-UJJ K28M-UJ3 K28M-UJ4 K28M-UJH K28M-UJ7 K28M-UJG K28M-UJD K28M-UJ2 K28M-UJB K28M-UJW K28M-UJ0 K28M-UJI K28M-UJX K28M-UJZ K28M-UJA K28M-UJC K28M-UJU K28M-UJ5 K28M-UJR K28M-UJV K28M-UJ1 K28M-UJ6 K28M-UJN K28M-UJE K28M-UJQ K28M-UJM K28M-UJS K28M-UJO K28M-UJT K28M-UJ9 K28M-UJL K28M-UJY K28M-UJP K28M-UJF|
|K28M-U38 K28M-U3K K28M-U3J K28M-U33 K28M-U34 K28M-U3H K28M-U37 K28M-U3G K28M-U3D K28M-U32 K28M-U3B K28M-U3W K28M-U30 K28M-U3I K28M-U3X K28M-U3Z K28M-U3A K28M-U3C K28M-U3U K28M-U35 K28M-U3R K28M-U3V K28M-U31 K28M-U36 K28M-U3N K28M-U3E K28M-U3Q K28M-U3M K28M-U3S K28M-U3O K28M-U3T K28M-U39 K28M-U3L K28M-U3Y K28M-U3P K28M-U3F|
© 2018 MissCitrus All Rights Reserved.

**Category-Aware Extreme-Tokenized Documents Filter**

Examples of low quality documents from base dataset FineWeb1.1.0 that our Category-Aware Extreme-Tokenized Documents Filter discards.

[Example 1: TokensPerChar = 0.527]

Peggy's Kitchen is a gourmet wedding cake and dessert bakery located in the beautiful city of San Diego. Peggy and I started this bakery with a dream of creating beautiful and tasty desserts. Within two years, we have grown from nobody to a well-known brand in the community. Many locals are drawn by our cakes and desserts, includes famous fashion blogger - Cubical Chic. If you ever had the chance to visit San Diego, don't forget to contact Peggy's Kitchen and order a cake or a fruit tart. It will be the highlight of your trip!
去年的這個時候因為P換工作，我們從聖地牙哥搬到矽谷。離開陽光沙灘海洋的南加州，一開始很不習慣。更不習慣的是要離開Peggy's Kitchen。Peggy's Kitchen 是我和Peggy一起創立的蛋糕甜點工作室。甜點研發跟製作大部分由Peggy一手掌控，我的工作則是幫甜點們拍出可口的照片和拍攝一些甜點製作的影片。其實更多的時間，我是負責「試吃」！Peggy's Kitchen 目前開業已兩年，並擁有忠實的客群。有機會去聖地牙哥的朋友們，不妨去嚐嚐Peggy的甜點，還有客製蛋糕的服務喔！
Peggy's Kitchen Facebook 粉絲頁

[Example 2: TokensPerChar = 0.519]

"My angel-faced Beloved holds the reins of the temporal and celestial worlds.
These two worlds are worth just a single strand of my Beloved's hair.
We cannot bear the allure of that gaze.
One rejuvenating glance would be enough for our lifetime.
Sometimes a sūfī[1], sometimes a zāhid[2], at others a qalandar[3];
Our unfathomable Beloved has many tints and shades.
Who, except the lover, would know the worth of [Beloved's] red gems?
But our eyes that shed pearls are aware of the value of rubies.

In the memory of [Beloved's] intoxicating eyes, Goya, with every breath;
Our wakeful hearts sip on the nectar of longing.
- A mystic
- Religious, devout, ascetic, perhaps suggestive of zealotry
- A wandering dervish

Dīn o dunyā dar kamand-i ān parī rukhsār-i mā
Har dō ālam qīmat-i yek tār-i muy-i yār-i mā
Mā nemī ārīm tāb-i ghamza-yi mizhgān-i ū
Yek nigāh-i jān fazāyash bas buvad dar kār-i mā
Gāh sūfī gāh zāhid gāh qalandar mī shavād
Rang hā-yi mukhtalif dārad but-i 'ayyār-i mā
Qadr-i l'al-i ū bajuz āshiq nādānad hīch kas
Qīmat-i yāqūt dānad chashm-i gohārbār-i mā
Har nafas guyā beh yād-i nargis-i makhmūr-i ū
Bādeh hā-yi shauq mī nushad dil-i hushyār-i mā

ਦੀਨੋ ਦੁਨੀਆ ਦਰ ਕਮੰਦਿ ਆਨ ਪਰੀ ਰੁਖਸਾਰਿ ਮਾ ।
ਹਰ ਦੋ ਆਲਮ ਕੀਮਤਿ ਯਕ ਤਾਰਿ ਮੁਇ ਯਾਰਿ ਮਾ ॥
ਮਾ ਨਮੀ ਆਰੀਮ ਤਾਬਿ ਗ਼ਮਸ਼ਾਹਇ ਮਿਜਗਾਨਿ ਉ ।
ਯਕ ਨਿਗਾਹਿ ਜਾਨ ਫ਼ਿਜ਼ਾਅਸ਼ ਬਸ ਬਵਦ ਦਰ ਕਾਰਿ ਮਾ ॥
ਗਾਹਿ ਸੂਫੀ ਗਾਹਿ ਜ਼ਾਹਦ ਗਹ ਕਲੰਦਰ ਮੀ ਸ਼ਵਦ ।
ਰੰਗਹਾਇ ਮੁਖਤਲਿਫ਼ ਦਾਰਦ ਬੁਤਿ ਅੱਯਾਰਿ ਮਾ ॥
ਕਦਰਿ ਲਾਲਿ ਉ ਬਜੁਜ਼ ਆਸ਼ਕ ਨਾਦਾਨਦ ਹੀਚ ਕਸ ।
ਕੀਮਤਿ ਯਾਕੂਤ ਦਾਨਦ ਚਸ਼ਮਿ ਗੋਹਾਰਬਾਰਿ ਮਾ ॥
ਹਰ ਨਫ਼ਸ ਗੋਯਾ ਬੇਹ ਯਾਦਿ ਨਰਗਸਿ ਮਖਮੂਰਿ ਉ ।
ਬਾਦੇਹ ਹਾਇ ਸ਼ੌਕ ਮੀ ਨੋਸ਼ਦ ਦਿਲਿ ਹੁਸ਼ਿਆਰਿ ਮਾ ॥

دین و دنیا در کمندِ آن پری رخسار ما
هر دو عالم قیمتِ یک تار موی یار ما
ما نمی آریم تاب غمزهٔ مژگانِ او
یک نگاهِ جان فزایش بس بود در کار ما
گاه صوفی گاه زاهد گه قلندر می شود
رنگ های مختلف دارد بت عیار ما
قدر لعل او بجز عاشق نداند هیچ کس
قیمتِ یاقوت داند چشمِ گوهربار ما
هر نفس گویا به یادِ نرگسِ مخمور او
باده های شوق می نوشد دلِ هشیار ما

The second ghazal from Bhai Nand Lal 'Goya' is an intimate exploration of Goya's relationship with the Guru. In his soaring first ghazal, Bhai Nand Lal offers a vivid account of his encounter with the Divine, which he describes as a stormy experience that brings him into the winds of reverence-bondage (bandigī). He describes a turn inward, a realization that while he is captured in the blue vault that is the sky, he can find freedom through constant remembrance of the Divine. He takes up his relationship with his Beloved in his second ghazal, which is both intimate in its details and vast in its love for the Guru, who holds reins of both the celestial and temporal realms (dīn ō dunīā).

In this ghazal, Goya describes an angel-faced Beloved whose perfection is that both the celestial and temporal realms are worth not even one strand of Beloved's hair. He offers a description of his Beloved's appearance: the lips that are red gems, the unbearable gaze. In the original Persian, the ghazal refers specifically to the flutter of the eyelashes of the Beloved, which we have simplified here for the sake of both brevity and clarity. The flutter of the eyelashes is so unbearable that even one glance from Beloved would sustain Goya in this lifetime.

In the last couplet, Goya metaphorizes his Beloved's intoxicating eyes as the narcissus flower (nargis), in whose memory he sips the nectar--or wine--of longing remembrance. The ghazal closing couplet brings to mind Puran Singh's understanding of simran as a state of "constant inebriation." This inebriated state is not a static one; it does not consist of the "dead peace" of the "Bhaktas of medieval India," for whom meditation entailed immersion into a "mystic reverie," a mindless state that "shuts itself up and shrivels up evidently in all ordinary practice to a mere

dead concept--all is one." Instead, this kind of simran causes one to become immersed in a "pool of nectar." This longing remembrance that brings one into a state of intoxication contemplates the "divine music of life;" it is a creative simran that necessitates "hard labor." This is perhaps the kind of simran Bhai Nand Lal is invoking as he takes every breath in memory of his Beloved's eyes.

The translators made several choices in translating the present ghazal that require some elaboration. First, we have chosen not to refer to the Beloved with gendered pronouns. Though most translations of classical Persian poetry would refer to the Beloved as female, we have chosen not to use gendered pronouns to refer to the Beloved as Bhai Nand Lal was writing in the court of and about Guru Gobind Singh Sahib. We found that by referring to the Beloved as such, without the mediation of pronouns, the translation is more precise and accessible for English-speaking readers who do not have a background in Persian poetry. Second, we have chosen not to translate sūfī, zāhid, or qalandar into English as it would not be possible to capture the meanings of these words in single English words. The ghazal text includes footnotes to which the reader can refer to understand this line better. We invite readers to engage in further research to develop their interpretation of this line of the ghazal."


[Example 3: TokensPerChar = 0.622]

"Archive for the 'Plutarch' Category
καὶ καθάπερ ὅταν ἐν συλλόγῳ τινὶ σιωπὴ γένηται, τὸν Ἑρμῆν ἐπεισεληλυθέναι λέγουσιν, οὕτως ὅταν εἰς συμπόσιον ἢ συνέδριον γνωρίμων λάλος εἰσέλθῃ, πάντες ἀποσιωπῶσι μὴ βουλόμενοι λαβὴν παρασχεῖν.
And just as, when a silence occurs in a meeting, they say 'Hermes has come in', so when a chatterbox comes in to a dinner-party or a gathering of friends, everyone falls silent, not wishing to let him get a hold.
The ancient equivalent of taking a deep breath and counting to ten.
Ἀθηνοδώρῳ δὲ τῷ φιλοσόφῳ διὰ γῆρας εἰς οἶκον ἀφεθῆναι δεηθέντι συνεχώρησεν. ἐπεὶ δὲ ἀσπασάμενος αὐτὸν ὁ Ἀθηνόδωρος εἶπεν, "ὅταν ὀργισθῇς, Καῖσαρ, μηδὲν εἴπῃς μηδὲ ποιήσῃς πρότερον ἢ τὰ εἴκοσι καὶ τέτταρα γράμματα διελθεῖν πρὸς ἑαυτόν," ἐπιλαβόμενος αὐτοῦ τῆς χειρός, "ἔτι σοῦ παρόντος," ἔφη, "χρείαν ἔχω", καὶ κατέσχεν αὐτὸν ἐνιαυτὸν ὅλον, εἰπὼν ὅτι "ἔστι καὶ σιγῆς ἀκίνδυνον γέρας."
He granted the request of the philosopher Athenodorus, who asked to be allowed to return home because of his old age. But when Athenodorus was taking his leave he said, 'Whenever you get angry, Caesar, say nothing and do nothing before you have run through the twenty-four letters of the alphabet to yourself.' Augustus seized hold of his hand and said, 'I still need you to be here!' and kept him for a whole year, saying 'The reward of silence is a lack of risk' [Simonides, fr. 582].
Plutarch, priest of Apollo at Delphi, doesn't really approve of Egyptian religion.
τοῦτο δ' οὐχ ἥκιστα πεπόνθασιν Αἰγύπτιοι περὶ τὰ τιμώμενα τῶν ζῴων. Ἕλληνες μὲν γὰρ ἔν γε τούτοις λέγουσιν ὀρθῶς καὶ νομίζουσιν ἱερὸν Ἀφροδίτης ζῷον εἶναι τὴν περιστερὰν καὶ τὸν δράκοντα τῆς Ἀθηνᾶς καὶ τὸν κόρακα τοῦ Ἀπόλλωνος καὶ τὸν κύνα τῆς Ἀρτέμιδος, ὡς Εὐριπίδης· "Ἑκάτης ἄγαλμα φωσφόρου κύων ἔσῃ"· Αἰγυπτίων δ' οἱ πολλοὶ θεραπεύοντες αὐτὰ τὰ ζῷα καὶ περιέποντες ὡς θεοὺς οὐ γέλωτος μόνον οὐδὲ χλευασμοῦ καταπεπλήκασι τὰς ἱερουργίας, ἀλλὰ τοῦτο τῆς ἀβελτερίας ἐλάχιστόν ἐστι κακόν· δόξα δ' ἐμφύεται δεινὴ τοὺς μὲν ἀσθενεῖς καὶ ἀκάκους εἰς ἄκρατον ὑπερείπουσα τὴν δεισιδαιμονίαν, τοῖς δὲ δριμυτέροις καὶ θρασυτέροις εἰς ἀθέους ἐμπίπτουσα καὶ θηριώδεις λογισμούς.
The Egyptians have fallen into no less an error in their worship of animals. For the Greeks speak of these matters in the correct way, and consider the dove to be the sacred animal of Aphrodite, the snake that of Athena, the raven that of Apollo, and the dog that of Artemis – as Euripides says: 'You shall be a dog, the image of Hecate the torch-bearer.' But most of the Egyptians do honour to the animals themselves and treat them with respect as though they were gods; not only have they filled the sacred rites with laughter and mockery – this is the smallest evil to come out of their silliness – but a terrible belief is implanted, which casts the weak and guileless into superstition and which brings down the more shrewd and bold into atheism and savage theorising.
περὶ δὲ τῶν Δημοσθένους λόγων ἐρωτηθείς, τίνα δοκοίη κάλλιστον εἶναι, τὸν μέγιστον εἶπε.
When he was asked which of Demosthenes' speeches he thought the best, he said, 'The longest one.'
It's the thought that counts.
Ἀρταξέρξης ὁ Περσῶν βασιλεύς, ὦ μέγιστε αὐτοκράτορ Καῖσαρ Τραϊανέ, οὐχ ἧττον οἰόμενος βασιλικὸν καὶ φιλάνθρωπον εἶναι τοῦ μεγάλα διδόναι τὸ μικρὰ λαμβάνειν εὐμενῶς καὶ προθύμως, ἐπεί, παρελαύνοντος αὐτοῦ καθ' ὁδόν, αὐτουργὸς ἄνθρωπος καὶ ἰδιώτης οὐδὲν ἔχων ἕτερον ἐκ τοῦ ποταμοῦ ταῖς χερσὶν ἀμφοτέραις ὕδωρ ὑπολαβὼν

προσήνεγκεν, ἡδέως ἐδέξατο καὶ ἐμειδίασε, τῇ προθυμίᾳ τοῦ διδόντος οὐ τῇ χρείᾳ τοῦ διδομένου τὴν χάριν μετρήσας.

Artaxerxes, the king of the Persians, o most high emperor Caesar Trajan, thought that receiving small gifts gladly and eagerly was no less regal and kindly to one's fellow-men than giving large gifts. When Artaxerxes was riding past on the road, a man who was a farmer, and just a member of the general public, took up water from the river (because he had nothing else) in his two hands and offered it to him; the king accepted it pleasantly and with a smile, measuring the favour by the giver's willingness rather than by the gift's usefulness.

χαρίεντος ἀνδρός, ὦ Σόσσιε Σενεκίων, καὶ φιλανθρώπου λόγον ἔχουσι Ῥωμαῖοι διὰ στόματος, ὅστις ἦν ὁ εἰπών, ἐπὶ μόνος ἐδείπνησεν, "βεβρωκέναι, μὴ δεδειπνηκέναι σήμερον", ὡς τοῦ δείπνου κοινωνίαν καὶ φιλοφροσύνην ἐφηδύνουσαν ἀεὶ ποθοῦντος.

Sossius Senecio, the Romans keep quoting the words of a charming and kind-hearted man who said, when he had dined alone, 'I have eaten, but I have not dined today' – since a dinner always needs sociability and friendliness as its seasoning.

ὁ μέντοι πρῶτος ἐκ τοῦ γένους Κικέρων ἐπονομασθεὶς ἄξιος λόγου δοκεῖ γενέσθαι διὸ τὴν ἐπίκλησιν οὐκ ἀπέρριψαν οἱ μετ' αὐτόν, ἀλλ' ἠσπάσαντο, καίπερ ὑπὸ πολλῶν χλευαζομένην. κίκερ γὰρ οἱ Λατῖνοι τὸν ἐρέβινθον καλοῦσι, κἀκεῖνος ἐν τῷ πέρατι τῆς ῥινὸς διαστολὴν ὡς ἔοικεν ἀμβλεῖαν εἶχεν ὥσπερ ἐρεβίνθου διαφυήν, ἀφ' ἧς ἐκτήσατο τὴν ἐπωνυμίαν. αὐτός γε μὴν Κικέρων, ὑπὲρ οὗ τάδε γέγραπται, τῶν φίλων αὐτὸν οἰομένων δεῖν, ὅτε πρῶτον ἀρχὴν μετῄει καὶ πολιτείας ἥπτετο, φυγεῖν τοὔνομα καὶ μεταθέσθαι, λέγεται νεανιευσάμενος εἰπεῖν, ὡς ἀγωνιεῖται τὸν Κικέρωνα τῶν Σκαύρων καὶ τῶν Κάτλων ἐνδοξότερον ἀποδεῖξαι.

The first member of the family who had the nickname 'Cicero' seems to have been worthy of note, because his descendants did not cast off the nickname, but were fond of it, even though it was ridiculed by many people. For Latin speakers call the chickpea 'cicer', and that ancestor, it seems, had a slight notch in the end of his nose, like the cleft in a chickpea, so from this he acquired the nickname. And when Cicero (the one about whom I am writing this biography) first began his public life and took up public office, his friends thought that he ought to drop or change his name, but he is said to have said, with youthful high spirits, that he would strive to make the name Cicero more renowned than Scaurus ['Bulging-ankles'] and Catulus ['Puppy']."



[Example 4: TokensPerChar = 0.599]

"Share the story of what Open Access means to you
University of Michigan needs your feedback to better understand how readers are using openly available ebooks. You can help by taking a short, privacy-friendly survey.
|ENHR||110.437 (Jun. 1995): 816-817||http://links.jstor.org/sici?sici=0013-8266%28199506%29110%3A437%3C816%3ASM%3E2.0.CO%3B2-O|
|AE||21.4 (Nov. 1994): 924-925||http://links.jstor.org/sici?sici=0094-0496%28199411%2921%3A4%3C924%3ASM%3E2.0.CO%3B2-G|
|Man||28.3 (Sep. 1993): 610-611||http://links.jstor.org/sici?sici=0025-1496%28199309%292%3A28%3A3%3C610%3ASM%3E2.0.CO%3B2-X|
|AATH||95.2 (Jun. 1993): 470-471||http://links.jstor.org/sici?sici=0002-7294%28199306%292%3A95%3A2%3C470%3ASM%3E2.0.CO%3B2-6|
1,776 views since June 25, 2018"



[Example 5: TokensPerChar = 1.116]

"ᐅᓂᒃᑲᖅᑐᐊᖃᑦᑕᓂᖅ ᓄᓇᕐᔪᐊᒥ ᓯᕗᓪᓕᖅᓯᓂᖕ ᑭᒡᒍᑎᖕᓄᑦ ᑲᔪᓯᑎᑕᐅᖃᑦᑕᖅᓯᒪᒪᑦ, ᐃᓄᖕᓂ ᐊᑐᐊᑎᑦᑎᓄᓂ ᐃᓕᖅᑯᓯᖕᑎᖕᓄᑦ ᓯᕗᓪᓕᖕᖓᓄᑦᓗ. ᐅᓂᒃᑐᖅᑲᐃᑦ ᐱᒻᒪᕆᐅᒻᒪᑦ ᐃᓄᐃᑦ ᐃᓕᖅᑯᓯᖕᑎᖕᓄᑦ. ᒫᓐᓇᐅᔪᖅ ᐅᑭᐅᖅᑕᖅᑐᒥ, ᑭᓯᐊᓂ, ᐅᓂᒃᑲᑦ ᑖᒃᑯᐊ ᑲᔪᓯᑎᑕᐅᖃᑕᖕᓂᒻᒪᑦ ᐊᓯᐅᑐᐃᓐᓇᕆᐊᖃᕐᒪᑦ."

ᕿᑭᖅᑕᓂ ᐃᓄᐃᑦ ᑲᑐᔾᔨᖃᑎᒌᒃᑯᑦ ᐃᓄᐃᑦ ᐃᓕᖅᑯᓯᕐᒥᓂᒃ ᓴᖅᐱᖅᑎᑦᑎᓇᓱᒃᐸᒃᐳᑦ ᓴᐳᒻᒥᔭᐅᓯᐊᖅᑐᑦᓗ. ᕿᑭᖅᑕᓂ ᐃᓄᐃᑦ ᑲᑐᔾᔨᖃᑎᒌᒃᑯᑦ ᓴᖅᑭᓚᐅᖅᐳᑦ Inuitmyths.com, ᐊᑐᐊᓇᐅᑎᑕᐅᓯᐊᖅᑐᑎᒃ ᐃᓕᖃᓂᒪᒋᔭᓂᒃ ᐃᓄᐃᑦ ᐅᓂᒃᖅᑐᐊᕐᒌᓂᒃ ᓄᓇᕗᒻᒥᐅᓄᑦ ᐃᓄᖕᓂᑦᓗ ᓄᓇᕐᔪᐊᕐᒥᐅᓂᑦ.

Inuitmyths.com ᕿᑭᖅᑕᓂ ᐃᓄᐃᑦ ᑲᑐᔾᔨᖃᑎᒌᒃᑯᑦ ᐱᓕᕆᐊᖃᐃᓐᓇᕐᓂᐊᖅᐳᑦ ᑲᑎᖅᓱᐃᓇᓱᖕᓂᒥᒃ ᐅᓂᒃᑐᖃᖅᓂᒃ ᐊᑐᐊᓇᐅᑎᓪᓗᒌᓪᓗ ᐃᓄᖕᓄᑦ. ᐅᓂᒃᖅᑲᕈᒪᔪᑦ ᑐᓴᖅᑕᐅᖁᔨᓂᒃ ᖃᐅᔨᒪᒍᕕᓪᓘᓐᓃᑦ ᐅᓂᒃᑕᓕᖕᓂᒃ, ᐅᖃᕐᑎᓄᑦ ᖃᐅᔨᒃᑲᐅᑎᒋᑦ ᐅᕗᖓ email@example.com. ᐱᓕᕆᖃᑎᒋᒃᑯᑦᑕ, ᖁᕕᐊᓱᒌᖃᑎᒋᓇᓱᖅᐳᒍᑦ ᓴᙱᒃᑎᒋᐊᕐᓗᑎᒍᓪᓗ ᐅᓂᒃᖅᑐᐊᕈᓯᖅᐳᑦ ᐱᒻᒪᕆᐅᒪᑦ ᐃᓄᐃᑦ ᐃᓕᖅᑯᓯᖕᒪᓂᑦ.

ᑲᑎᖅᓱᐃᓂᖅ ᐅᓂᒃᖃᓂᒃ ᑲᑐᔾᔨᖃᑎᒌᖕᓂᐅᕘᖅ. ᕿᑭᖅᑕᓂ ᐃᓄᐃᑦ ᑲᑐᔾᔨᖃᑎᒌᒃᑯᑦ ᖁᔭᓇᕈᒪᕗᑦ ᐱᓕᕆᖃᑎᖃᖅᖃᕐᓂᒥᒃ ᐃᒃᔭᖅᓱᒪᔪᒥᒃ ᐅᕙᑦᑎᓂᒃ.

ᓄᓇᕗᒻᒥ ᒪᕐᕉᖕᓂᒃ ᐅᖃᐅᓯᓕᖕᓂᑦ ᐃᓕᓐᓂᐊᖅᑎᑦᑎᓂᕐᒧᑦ ᑲᑐᔾᔨᖃᑎᒌᖕᒌᑦ

ᓄᓇᕗᒻᒥ ᐃᓕᓴᐃᔨᖅᓴᐃᑦ ᐃᓕᓐᓂᐊᖅᐸᓕᔅ

ᑲᓇᑕᒥ ᑐᓴᕋᔅᓴᓕᕆᔨᒃᑯᑦ ᑎᒥᖕᒥᒃ

Storytelling traditions around the world are passed from generation to generation, linking people to their cultures and ancestors. Traditional stories are an important aspect of Inuit culture. Currently in the Arctic, however, many of these stories are not being passed on and are at risk of being lost.

The Qikiqtani Inuit Association (QIA) works hard to promote and protect Inuit culture. QIA has developed Inuitmyths.com, to provide a resource for Nunavummiut and people from around the world who want to learn more about the Inuit storytelling tradition.

Inuitmyths.com is QIA's ongoing initiative to collect traditional stories and make them available to the public. If you have stories you would like to share or if you know someone who does, please contact us at firstname.lastname@example.org. By working together, we will be able to celebrate and strengthen our storytelling tradition as an integral part of Inuit culture.

Collecting these stories is a shared effort. QIA wishes to thank our collaborative partners who have assisted us.
Our project partners are:
Nunavut Bilingual Education Society (NBES)
Nunavut Teacher Education Program (NTEP)
Nunavut Arctic College (NAC)
Department of Culture, Elders, Language and Youth (CLEY)
Department of Education
Canadian Broadcasting Corporation (CBC)"



\end{document}